\documentclass[12pt]{article}

\usepackage[a4paper,top=3cm,bottom=2cm,left=2cm,right=2cm,marginparwidth=1.25cm]{geometry}

\usepackage{adjustbox}
\usepackage{microtype}
\usepackage{graphicx}
\usepackage{subfigure}
\usepackage{booktabs} % for professional tables
\usepackage[english]{babel}
\usepackage[utf8x]{inputenc}
\usepackage[T1]{fontenc}
\usepackage{comment}
\usepackage{booktabs}
\newcommand{\rev}[1]{{\color{black}{#1}}}
\newcommand{\beginsupplement}{%
        \setcounter{table}{0}
        \renewcommand{\thetable}{S\arabic{table}}%
        \setcounter{figure}{0}
        \renewcommand{\thefigure}{S\arabic{figure}}         \setcounter{section}{0}
        \renewcommand{\thesection}{S\arabic{section}}
}
\usepackage{amsmath}
\usepackage{graphicx}
\usepackage{amsmath}
\usepackage{amsthm}
\usepackage{amssymb}
\usepackage{bbm}
\usepackage{verbatim}
\usepackage{graphicx}
\usepackage{subfigure}
\usepackage{afterpage}
\usepackage{etoolbox}
\usepackage{float}
\usepackage{rotating}
\usepackage[inline]{enumitem}

\usepackage{algorithmic}
\usepackage{algorithm}% http://ctan.org/pkg/algorithm
\usepackage[colorinlistoftodos]{todonotes}
\usepackage[colorlinks=true, allcolors=blue]{hyperref}
\newtheorem{props}{Proposition}
\newcommand {\myvec}[1] {{\mbox{\boldmath $#1$}}}
\newcommand{\myx}{\myvec{x}}
\newcommand{\myz}{\myvec{z}}
\newcommand{\myth}{\myvec{\theta}}
\newcommand{\mys}{\myvec{s}}
\newcommand{\tils}{\tilde{S}}
\newcommand{\prob}{\mathbb{P}}

\usepackage[colorinlistoftodos]{todonotes}

\newcommand{\exval}{\mathbb{E}}
\newcommand{\norm}[1]{\left\lVert#1\right\rVert}
\title{Feature selection using Stochastic Gates }
\author{Yutaro Yamada $^{1 \ast }$ \and Ofir Lindenbaum $^{2 \ast }$ \and Sahand Negahban$^{1}$ \and Yuval Kluger$^{2,3,4 \dagger}$\\
\\
\normalsize{$^{1}$Statistics Department;} 
\normalsize{$^{2}$Applied Mathematics Program;}\\
\normalsize{$^{3}$Computational Biology and Bioinformatics;}\\
\normalsize{$^{4}$Department of Pathology;}\\
\normalsize{Yale University, New Haven, CT, USA}\\
\\
\normalsize{$^\dagger$Corresponding author. E-mail: yuval.kluger@yale.edu}\\\normalsize{Address: 333 Cedar St, New Haven, CT 06510, USA}\\
\normalsize{$^\ast$ These authors contributed equally.}
}
\date{}
\begin{document}
\maketitle

\begin{abstract}
Feature selection problems have been extensively studied in the setting of linear estimation (e.g. LASSO), but less emphasis has been placed on feature selection for non-linear functions. 
In this study, we propose a method for feature selection in neural network estimation problems. The new procedure is based on probabilistic relaxation of the $\ell_0$ norm of features, or the count of the number of selected features. Our $\ell_0$-based regularization relies on a continuous relaxation of the Bernoulli distribution; such relaxation allows our model to learn the parameters of the approximate Bernoulli distributions via gradient descent. The proposed framework simultaneously learns either a nonlinear regression or classification function while selecting a small subset of features. We provide an information-theoretic justification for incorporating Bernoulli distribution into feature selection. Furthermore, we evaluate our method using synthetic and real-life data to demonstrate that our approach outperforms other commonly used methods in both predictive performance and feature selection.
\end{abstract}

\section{Introduction}

Feature selection is a fundamental task in machine learning and statistics. Selecting a subset of relevant features may result in several potential benefits: reducing experimental costs \cite{min2014feature}, enhancing interpretability \cite{Ribeiro:2016:WIT:2939672.2939778}, speeding up computation, reducing memory and even improving model generalization on unseen data \cite{survey}. 
For example, in biomedical studies, machine learning can provide effective diagnostics or prognostics models. 
However, the number of features (e.g., genes or proteins) often exceeds the number of samples. 
In this setting, feature selection can lead to improved risk assessment and provide meaningful biological insights.
While neural networks are good candidates for learning diagnostics models, identifying relevant features while building compact predictive models remains an open challenge.

\begin{figure*}[htb!]
\vskip -0.12in
\begin{center}
{\includegraphics[width=0.48\textwidth] {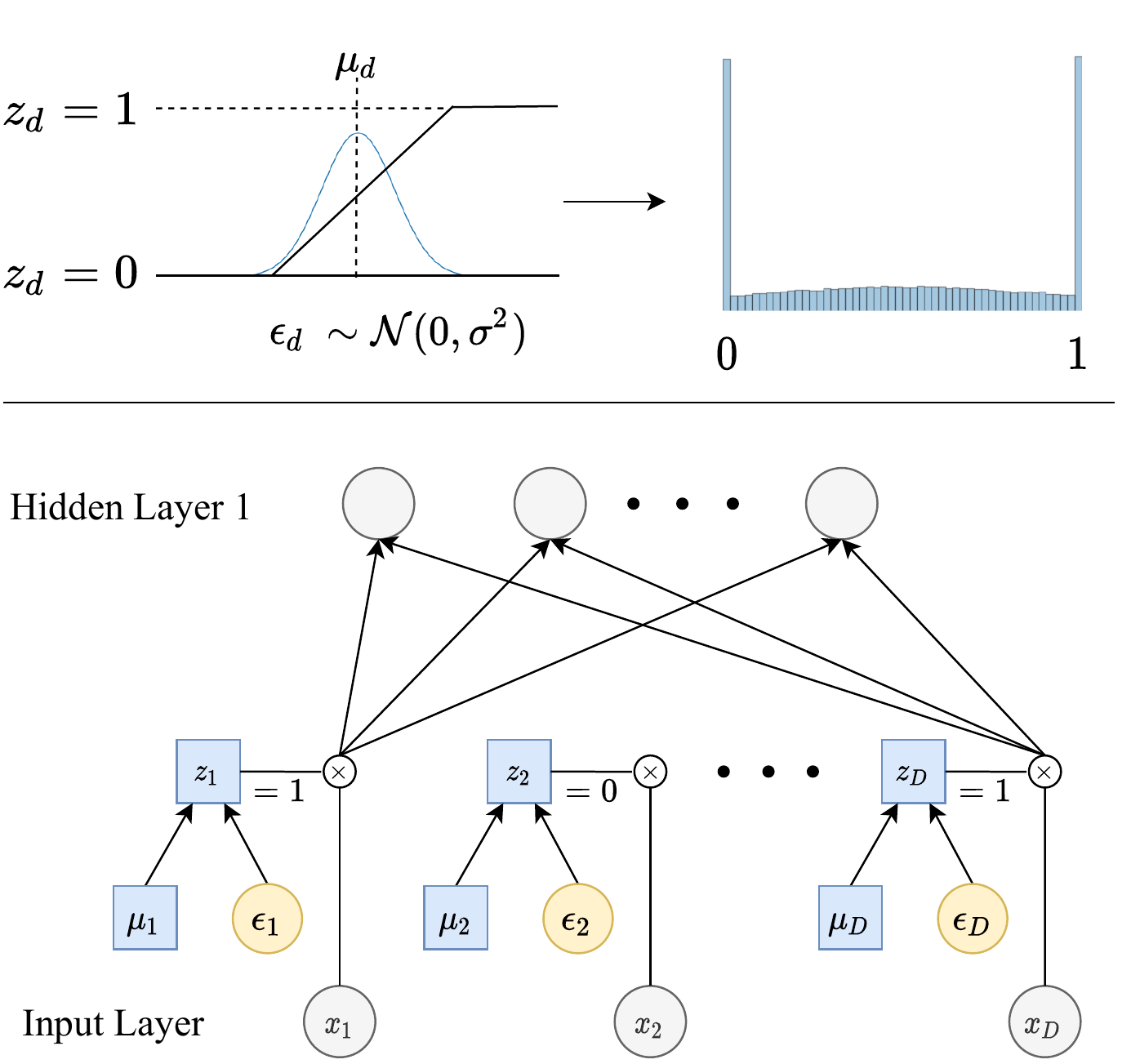}}
{\includegraphics[width=0.48\textwidth] {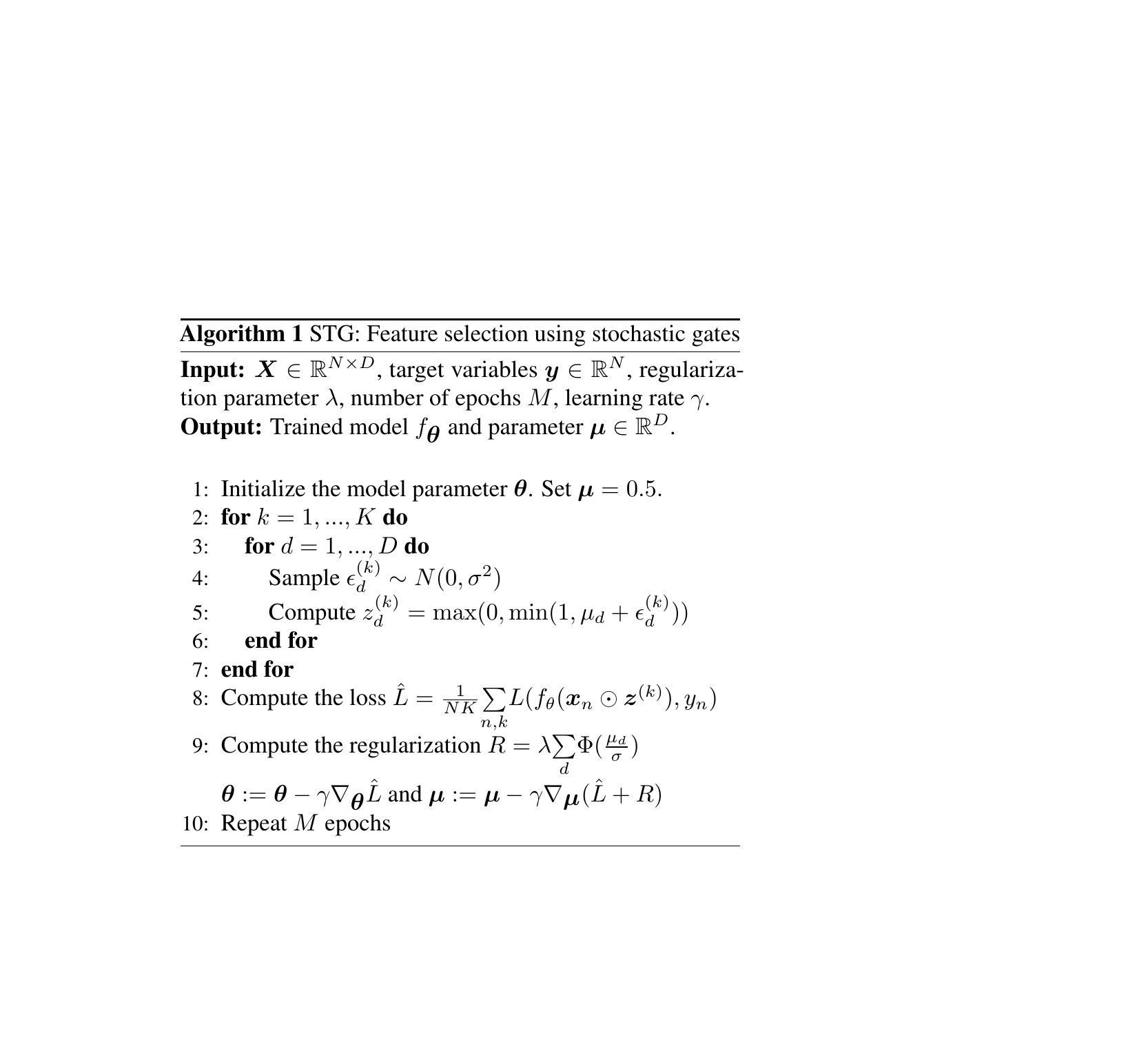}}
\end{center}
\vskip -0.15in
\caption{Top left: Each stochastic gate $z_d$ is drawn from the STG approximation of the Bernoulli distribution (shown as the blue histogram on the right). Specifically, $z_d$  is obtained by applying the hard-sigmoid function to a mean-shifted Gaussian random variable (step 5 in algorithm 1). Bottom left: The $z_d$ stochastic gate is attached to the $x_d$ input feature, where the trainable parameter $\mu_d$ controls the probability of the gate being active. Right: Pseudocode of our algorithm for feature selection. See the supplementary material for a discussion of $\sigma$ and $\lambda$'s selection.
} 
\vskip -0.26in
\label{fig:illustrate}
\end{figure*}

Feature selection methods are classified into three major categories: filter methods, wrapper methods, and embedded methods. \textbf{Filter methods} attempt to remove irrelevant features prior to learning a model. These methods filter features using a per-feature relevance score that is created based on statistical measures \cite{MI1,MI2,MI3,HSIC1,HSIC2,HSIC3}. Wrapper methods \cite{wrapper1,wrapper2,wrapper3,wrapper4,KernelW} use the outcome of a model to determine the relevance of each feature. \textbf{Wrapper methods} require recomputing the model for each subset of features and, thus, become computationally expensive, especially in the context of deep neural networks \cite{DNN1,DNN2,DNN3}. \textbf{Embedded methods} aim to remove this burden by learning the model while simultaneously selecting the subset of relevant features. 
The Least Absolute Shrinkage and Selection Operator (LASSO) \cite{Lasso} is a well-known embedded method, whose objective is to minimize the loss while enforcing an $\ell_1$ constraint on the weights of the features. LASSO is scalable and widely used \cite{Lasso2,Lasso3,Lasso4}, but it is
restricted to the domain of linear functions and suffers from shrinkage of the model parameters. It seems natural to extend the LASSO using neural networks; however, gradient descent on an $\ell_1$ regularized objective neither performs well in practice nor sparsifies the input layer \cite{DFS,sparseNN-group-lasso,one-layer}.

To overcome these limitations, we develop a fully \textit{embedded feature selection} method for nonlinear models. 
Our method improves upon the LASSO formulation by: a) capturing nonlinear interactions between features via neural network modeling and b) employing an $\ell_0$-like regularization using gates with weights parametrized by a smooth variant of a Bernoulli distribution. These two improvements are jointly formulated as a fully differentiable neural network that provides a solution to the important long-standing problem of feature selection for nonlinear functions. 

%\rev{The figure 1 drawing doesn't look right. The probabilities should be concentrated in the center but they're concentrated on the outskirts.}

Specifically, our contributions are as follows:
 \begin{itemize}

    \item We identify the limitations of the logistic-distribution-based Bernoulli relaxation \cite{Gumbel2,Gumbel1,Louizos2017LearningSN} in feature selection and present a Gaussian-based alternative termed stochastic gate (STG), which is better in terms of model performance and consistency of feature selection.

    \item We develop an embedded nonlinear feature selection method by introducing the stochastic gates to the input layer (the feature space) of a neural network. 
    \item We justify our probabilistic approach by analyzing the constrained Mutual Information maximization objective of feature selection.

\end{itemize}
We demonstrate the advantages of our method for classification, regression, and survival analysis tasks using numerous examples.

\paragraph{Notation:}
Vectors are denoted by {\textbf{bold}} lowercase letters $\myvec{x}$ and random vectors as bold
uppercase letters $\myvec{X}$. Scalars are denoted by lower case letters $y$, while random
variables are uppercase $Y$. A set is represented by a script font 
$\cal{S}$. For example the $n^{th}$ vector-valued observation is denoted as
$\myvec{x}_n$ whereas $X_d$ represents the $d^{th}$ feature of the vector-valued
random variable $\myvec{X}$. Let $[n] = {1,2,\hdots,n}$. For a set $\cal{S}
\subset [D]$ let the vector $\mys \in \{0,1\}^D$ be the characteristic function for the set.
That is $s_i = 1$ if $i \in \cal{S}$ and $0$ otherwise. For two vectors $\myvec{x}$ and $\myvec{z}$ we denote $\myx \odot \myz$ to be the element-wise product between $\myx$ and
$\myz$. Thus, if we let $\mys \in \{0,1\}^D$ be the characteristic vector of
$\cal{S}$, then  we may define $\myx_{\cal{S}} = \myx \odot \mys$. The $\ell_1$
norm of $\myx$ is denoted by $\|\myx\|_1 = \sum_{i=1}^D |x_i|$. Finally, the
$\ell_0$ norm of $\myx$ is denoted by $\|\myx\|_0$ and counts the total number
of non-zero entries in the vector $\myx$.

\section{Problem Setup and Background}
\label{gen_inst}

Let $\mathcal{X} \subset \mathbb{R}^D$ be the input domain with corresponding
response domain $\mathcal{Y}$. Given realizations from some unknown data
distribution $P_{{X},{Y}},$ the goal of embedded feature selection methods is to
simultaneously select a subset of indices ${\cal{S}} \subset \{1,...D\}$ and construct a model $f_{\myvec{\theta}} \in \mathcal{F}$ that predicts $Y$ based on the selected features $\myvec{X}_{\cal{S}}$. 

Given a loss $L$, the selection of features $\cal{S} \subset [D]$, and choice
of parameters $\myth$ can be evaluated in terms of the following risk: 
\begin{align} \label{eq:optim}
    R(\myvec{\theta}, \myvec{s}) = \mathbb{E}_{{X},{Y}} L(f_{\theta}(\myvec{X \odot s}), {Y}) , 
\end{align}
where we recall that $\myvec{s} = \{0, 1\}^D$ is a vector of indicator variables
for the set $\cal{S}$, and $\odot$ denotes the point-wise product. Embedded feature selection methods search for parameters $\myth$ and $\mys$ that minimize $R(\myth,\mys)$ such
that $\|\mys\|_0$ is small compared to $D$. 

\subsection{Feature Selection for Linear Models}

We first review the feature selection problem in the linear setting for a least squares loss. 
Given observations $\{\myvec{x}_n, {y}_n\}_{n=1}^N$, a natural objective derived from \eqref{eq:optim} is the
constrained empirical risk minimization  
\begin{equation} \label{eq:1}
    \min_{\myvec{\theta}} \frac{1}{N} \sum_{n=1}^N (\myvec{\theta}^T \myvec{x}_n - y_n)^2 \quad \textrm{s.t. $\|\myvec{\theta}\|_0 \leq k$}.
\end{equation}
Since the above problem is intractable, several authors replace the $\ell_0$
constraint with a surrogate function, $\Omega(\myvec{\theta}):\mathbb{R}^D
\rightarrow \mathbb{R}_{+}$, designed to penalize the number of selected features
in $\myth$. A popular choice for $\Omega$ is the $\ell_1$ norm, which yields a
convex problem and more precisely the LASSO optimization~\cite{Lasso}.
Computationally efficient
algorithms for solving the LASSO problem have been proposed~\cite{Lasso, nesterov2013gradient,qian2019fast}. While
the original LASSO focuses on the constrained optimization problem, the
regularized least squares formulation, which is often used in practice, yields the
following minimization objective: 
\begin{equation}
\label{eq:lasso}
   \min_{\myth} \frac{1}{N} \sum_{n=1}^N  (\myvec{\theta}^T\myvec{x}_n - y_n)^2 + \lambda \| \myvec{\theta} \|_1.
\end{equation} 
The hyperparameter $\lambda$ trades off the amount of regularization versus the
fit of the objective\footnote{$\lambda$ has a one-to-one correspondence to $k$ in the convex setting via Lagrangian duality.}. The $\ell_1$-regularized method is
effective for feature selection and prediction; however, it achieves this
through shrinkage of the coefficients and is restricted to linear models. To avoid shrinkage, non-convex choices for $\Omega$ have been proposed~\cite{fan2001variable}. As demonstrated in several studies~\cite{nonconvex0,nonconvex1,nonconvex2}, non-convex regularizers perform well
both theoretically and empirically in prediction and feature selection.

Our goal is to develop a regularization technique that both avoids shrinkage and performs feature selection while learning a nonlinear function. To allow nonlinearities, Kernel methods have been
considered~\cite{yamada2}, but scale quadratically in the number of
observations. An alternative approach is to model $f_{\myth}$ using a neural network with $\ell_1$ regularization on the input weights~\cite{DFS,sparseNN-group-lasso,one-layer}. However, in practice, introducing an $\ell_1$ penalty into gradient descent does not sparsify the weights and requires post-training thresholding. Below, we present our method that applies a differentiable approximation of an $\ell_0$ penalty on the first layer of a neural network.

\section{Proposed Method}
\label{sec:pro}

To implement an $\ell_0$ regularization to either linear or nonlinear models, we introduce a probabilistic and computationally efficient neural network approach. 
It is well known that an exact $\ell_0$ regularization is computationally expensive and intractable for high dimensions. Moreover, the $\ell_0$ norm cannot be incorporated into a gradient descent based optimization. To overcome these limitations, a probabilistic formulation provides a compelling alternative. Specifically, we introduce Bernoulli gates applied to each of the $d$ input nodes of a neural network. A random vector $\myvec{\tilde{S}}$ represents these Bernoulli gates, whose entries are independent and
satisfy $ \prob(\tils_d = 1)=\pi_d$ for $d \in [D]$, respectively. If we denote the
empirical expectation over the observations as $\hat{\mathbb{E}}_{X,Y}$, then, the empirical regularized risk (Eq.~\ref{eq:optim}) becomes
\begin{equation}
\label{eq:bern_risk}
   \hat{R}(\myvec{\theta}, \myvec{\pi}) =  \hat{\mathbb{E}}_{{X},{Y} } \mathbb{E}_{\tilde{S}} \left [ L(f_{\theta}(\myvec{X} \odot \myvec{\tilde{S}}), Y) + \lambda  ||\myvec{\tilde{S}}||_0  \right ] ,
\end{equation} 
where $\mathbb{E}_{\tilde{S}}||\myvec{\tilde{S}}||_0$ boils down to the sum of Bernoulli parameters $ \sum_{d=1}^D
\pi_d$. Note that, if we constrain $\pi_d \in \{0,1\}$, this formulation is
equivalent to the constrained version of equation~\eqref{eq:optim}, with a regularized penalty on cardinality rather than an explicit constraint. Moreover, this probabilistic formulation converts the combinatorial search to a search over the space of Bernoulli distribution parameters (also motivated in Section \ref{sec:MI}). Thus, the problem of feature selection translates to finding $\myvec{\theta}^*$ and $\myvec{\pi}^*$ that minimize the empirical risk based on the formulation in Eq. \ref{eq:bern_risk}.

Minimization of the empirical risk via gradient descent seems like a natural way to simultaneously determine the model parameters $\myvec{\theta}^*$ and Bernoulli-based feature selection parameters $\myvec{\pi}^*$. However, optimization of a loss function, which includes discrete random variables, suffers from high variance (see supplementary for more details and \cite{mnih2016variational}). To overcome this limitation, several authors have proposed using a continuous approximation of discrete random variables, such as the Concrete \cite{Gumbel1, Gumbel2} or Hard-Concrete (HC) \cite{Louizos2017LearningSN}.

We observed that the HC still suffers from high variance and, thus, is not suited for the task of feature selection. Therefore, we develop an empirically superior continuous distribution that is fully differentiable and implemented only to activate or deactivate the gates linking each feature (node) to the rest of the network. Our method provides an embedded feature selection algorithm with superior results in terms of both accuracy and capturing informative features compared with the state-of-the-art.  

\subsection{Bernoulli Continuous Relaxation for Feature Selection}
\label{sec:cr}
Feature selection requires stability in the selected set of features. 
The use of logistic distributions such as the Concrete \cite{Gumbel1,Gumbel2} and HC \cite{Louizos2017LearningSN} induces high variance in the approximated Bernoulli variables due to the heavy-tailedness, which often leads to inconsistency in the set of selected features. 
To address such limitations, we propose a Gaussian-based continuous relaxation for the Bernoulli variables $\tilde{S}_d$ for $d \in [D]$. 
We refer to each relaxed Bernoulli variable as a stochastic gate (STG) defined by $z_d = \max(0,
 \min(1, \mu_d + \epsilon_d))$, where $\epsilon_d$ is drawn from  $\mathcal{N}(0 ,\sigma^2)$ and $\sigma$ is fixed throughout training. This approximation
 can be viewed as a clipped, mean-shifted, Gaussian random variable as shown in the left part of Fig. \ref{fig:illustrate}. 
Furthermore, the gradient of the objective with respect to $\mu_d$ can be computed via the chain rule, which is commonly known as the reparameterization trick \cite{reparameterization1,reparameterization2}.

We can now write our objective as a minimization of the empirical risk $\hat{R}(\myvec{\theta}, \myvec{\mu})$:
\begin{equation} \label{eq:risk_p}
  \min_{\myvec{\theta, \mu}}\hat{\mathbb{E}}_{{X},{Y} } \mathbb{E}_{{{Z}}} \left [ L(f_{\theta}(\myvec{{X}}\odot \myvec{Z}), {Y}) + \lambda  ||\myvec{{Z}}||_0  \right ],
\end{equation} where $\myvec{Z}$ is a random vector with $D$ independent variables $z_d$ for $d \in [D]$. Under the continuous relaxation, the expected regularization term in the
objective $\hat{R}(\myvec{\theta}, \myvec{\mu})$ (Eq. \ref{eq:risk_p}) is simply the sum of the
probabilities that the gates $\{z_d\}_{d=1}^D$ are active \rev{or $\sum_{d \in [D]} \prob(z_d > 0)$}. This sum is equal to
$\sum_{d=1}^D \Phi \left ( \frac{\mu_d}{\sigma} \right)$, where
$\Phi$ is the standard Gaussian CDF. To optimize the empirical surrogate of the objective (Eq. \ref{eq:risk_p}), we first differentiate it with respect to $\myvec{\mu}$. This computation is done using a Monte Carlo sampling gradient estimator which gives
\begin{align*}
    \frac{1}{K} \sum_{k=1}^K \left [ L'(\myvec{{z}}^{(k)}) \frac{\partial {z}^{(k)}_d}{\partial \mu_d}  \right ] + \rev{\lambda \frac{\partial}{\partial \mu_d}  \Phi \left ( \frac{\mu_d}{\sigma} \right)} , 
\end{align*} where $K$ is the number of Monte Carlo samples. Thus, we can update
the parameters $\mu_d$ for \rev{$d \in [D]$} via gradient descent.

 Altogether, Eq. \ref{eq:risk_p} is optimized using SGD over the model parameters $\myvec{\theta}$ and the parameters $\myvec{\mu}$, where the latter substitute the parameters $\myvec{\pi}$ in Eq. \ref{eq:bern_risk}. See Algorithm 1 for a pseudocode of this procedure.

To remove the stochasticity from the learned gates after training, we set $\hat{z}_d = \max(0, \min(1, \mu_d))$, which informs what features are selected. 
\rev{In our experiments, for all synthetic datasets, we observe that the coordinates of $\hat{z}$ converge to 0 or 1.
However, when the signal is weak (e.g. the class samples are not separated) training the gates until convergence may cause overfitting of the model parameters. In these cases, setting a cutoff value (e.g. 0.5) and performing early stopping is beneficial.} 
%\del{Note that when $0 < \mu_d < 1$, $\hat{z}_d$ returns the
%value between $(0, 1)$. In such a case, we can treat the value of $\hat{z}_d$ as
%feature importance or employ an additional thresholding (i.e. 1 if $\hat{z}_d >
%0.5$ and $0$ otherwise) depending on application-specific needs.} 
In the supplementary material, we discuss our choice of $\sigma$.

\section{Connection to Mutual Information}
\label{sec:MI}
In this section, we use a Mutual Information (MI) perspective to show an equivalence between a constrained $\ell_0$-based optimization for feature selection and an optimization over Bernoulli distribution parameters.

\subsection{Mutual Information based objective}
From an information theoretic standpoint, the goal of feature selection is to find
the subset of features $\cal{S}$ that has the highest Mutual Information (MI) with
the target variable $Y$. MI between two random variables can be
defined as  $I(\myvec{X}; {Y}) = H({Y}) -
H({Y}|\myvec{X})$, where $H({Y})$ and $H({Y}|\myvec{X})$ are the entropy of
$p_{{Y}}({Y})$ and the conditional entropy of $p_{{Y}|\bf{X}}({Y}|\myvec{X})$,
respectively \cite{Cover:2006:EIT:1146355}. We can then formulate the task as
selecting $\cal{S}$ such that the mutual information between $\myvec{X}_{\cal{S}}$
and $Y$ is maximized:

\begin{equation}\label{eq:MI_c}
\max_{{\cal{S}} } I(\myvec{X}_{\cal{S}}; Y) \quad \text{s.t. } \lvert {\cal{S}} \rvert = k,
\end{equation}
where $k$ is the hypothesized number of relevant features. 

\begin{figure*}[htb!]
\vskip -0.1in
\begin{center}
\includegraphics[width=0.32\textwidth] {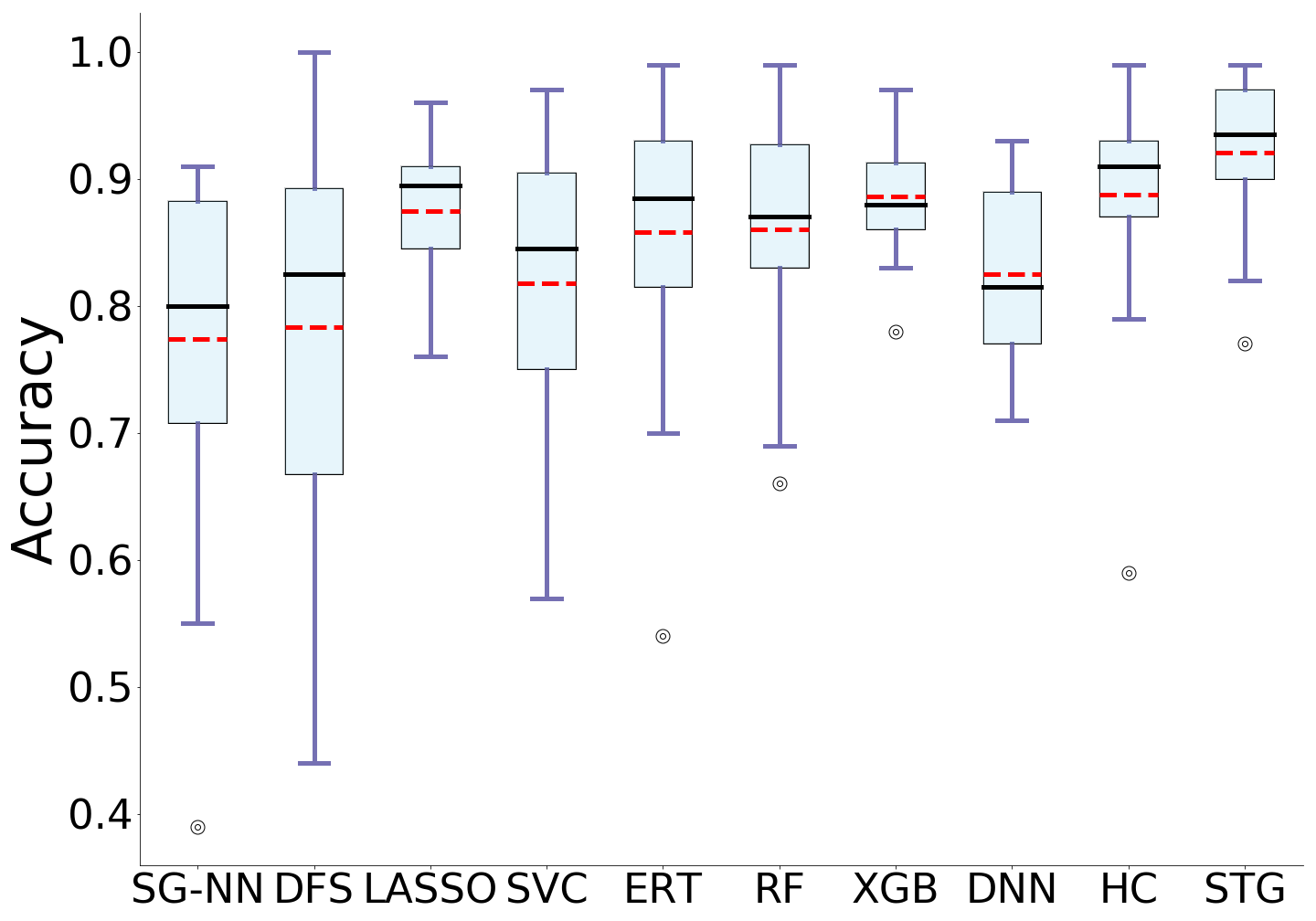}
\includegraphics[width=0.32\textwidth] {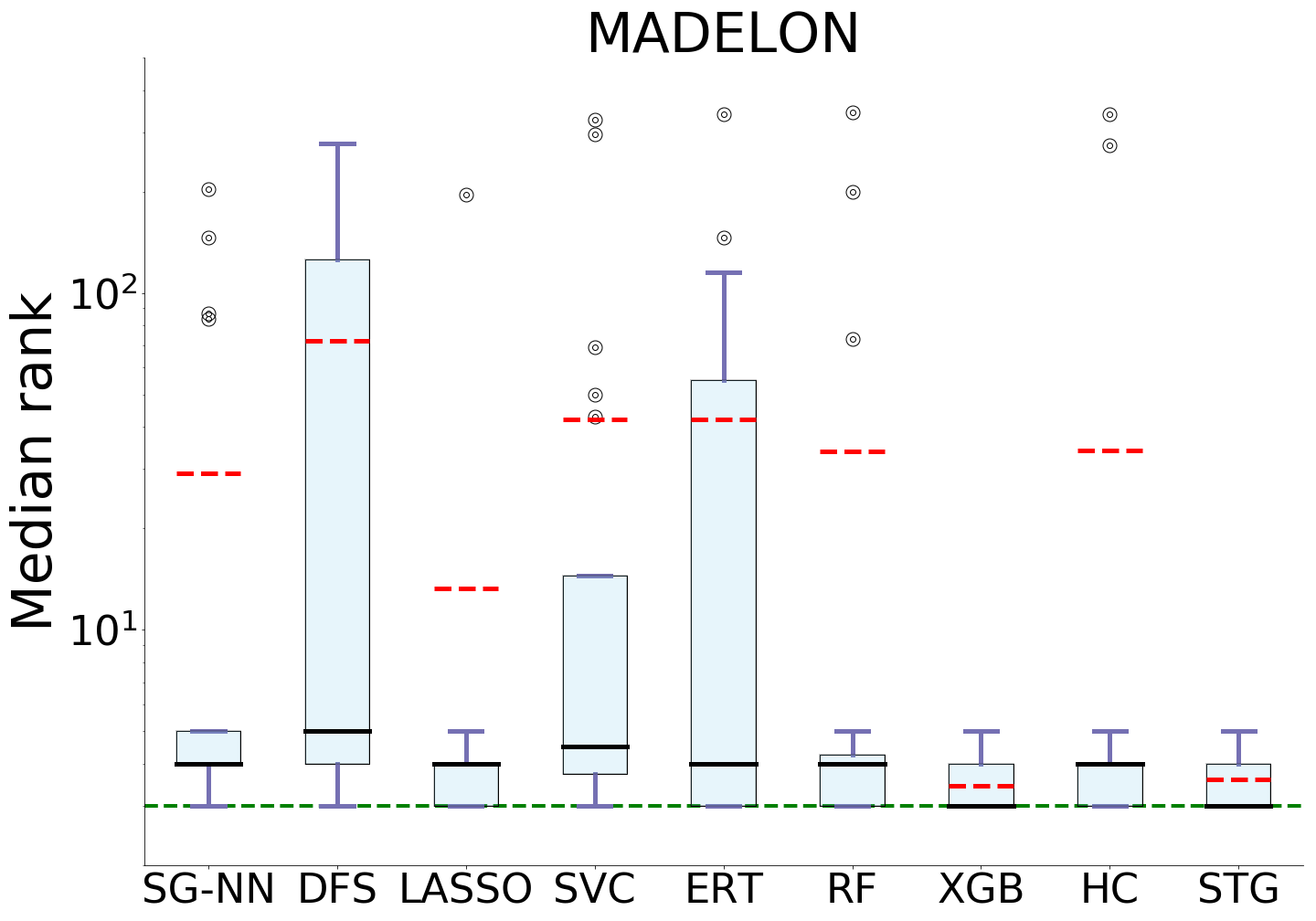}
\includegraphics[width=0.32\textwidth] {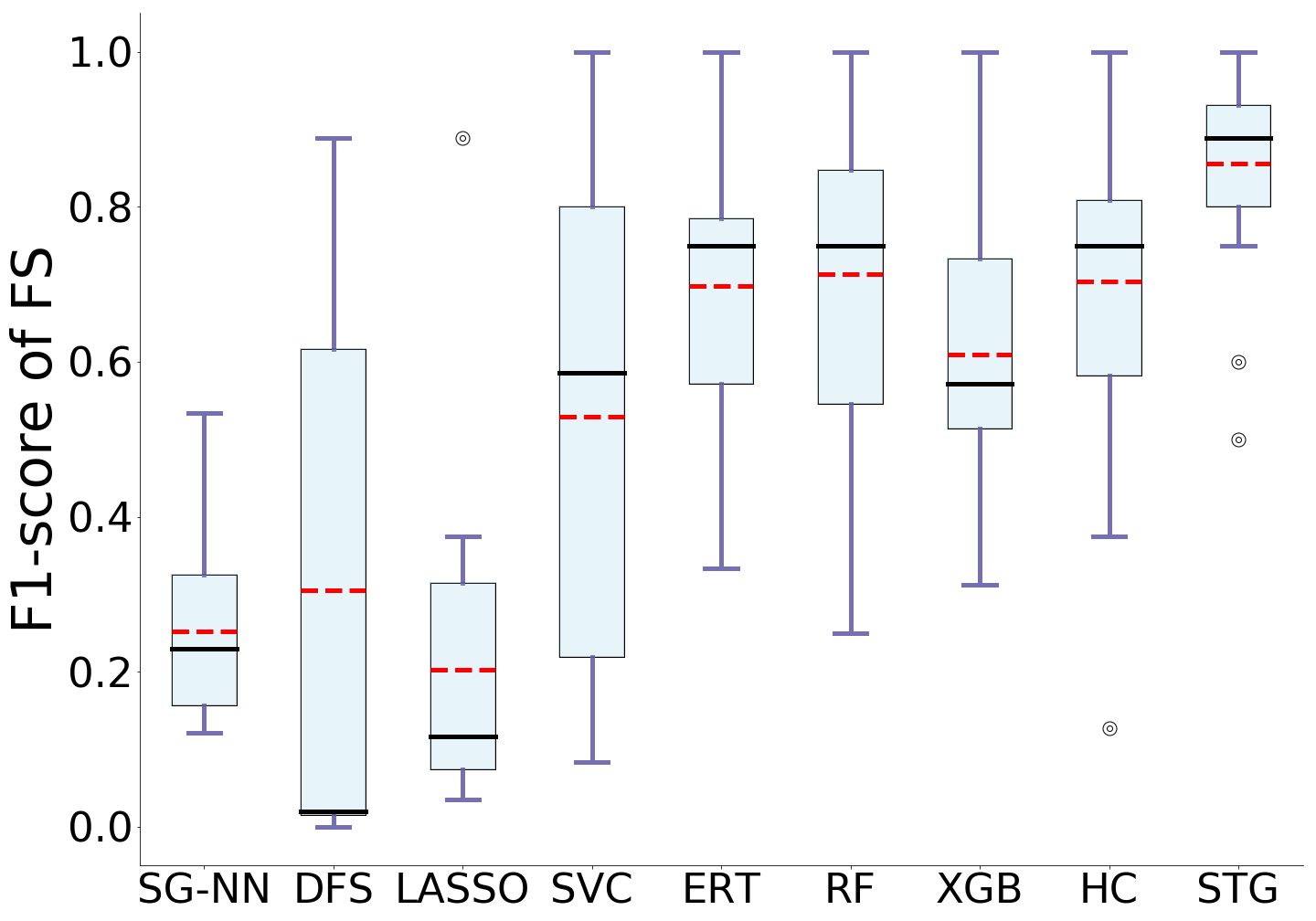}
\includegraphics[width=0.32\textwidth] {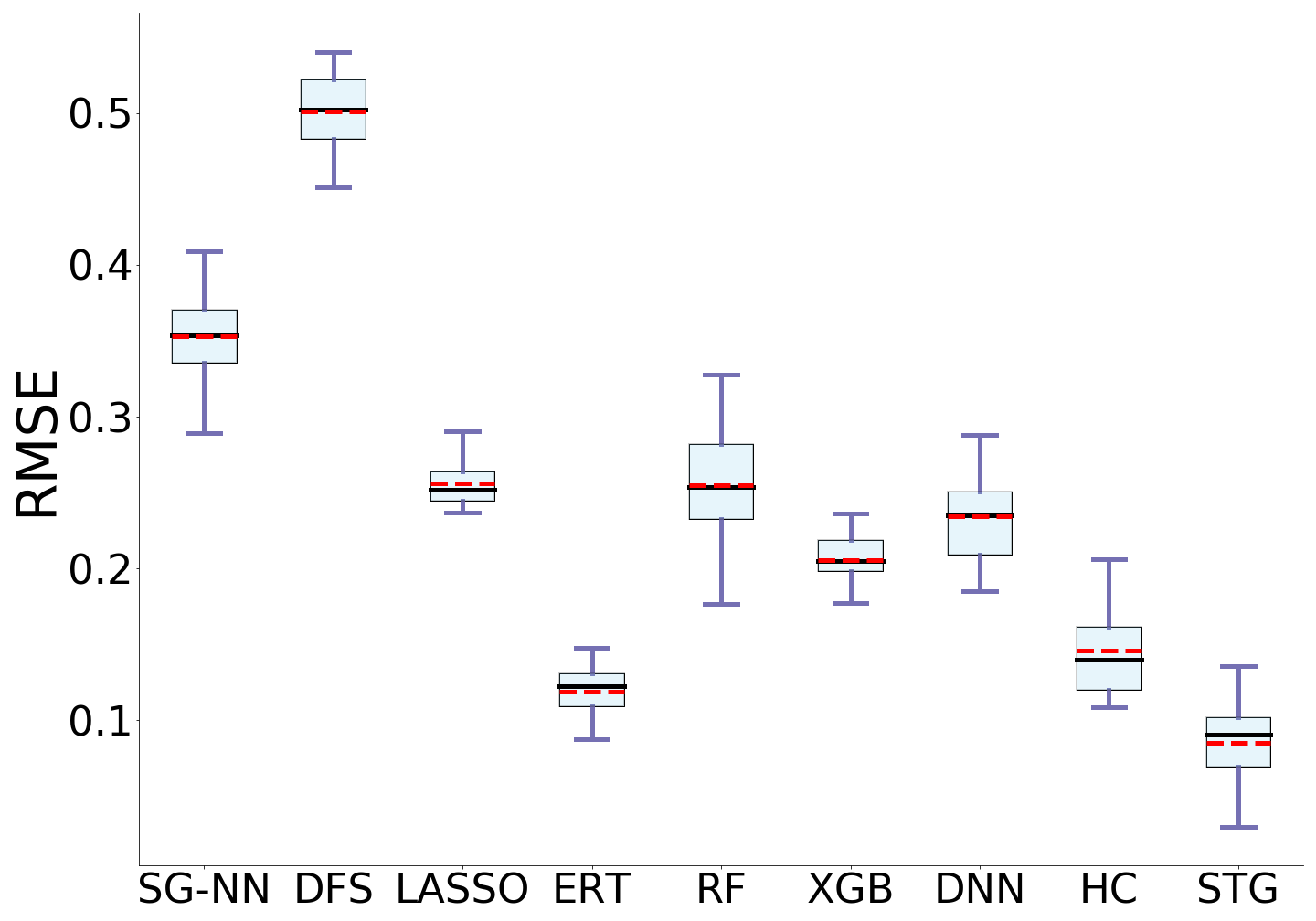}
\includegraphics[width=0.32\textwidth] {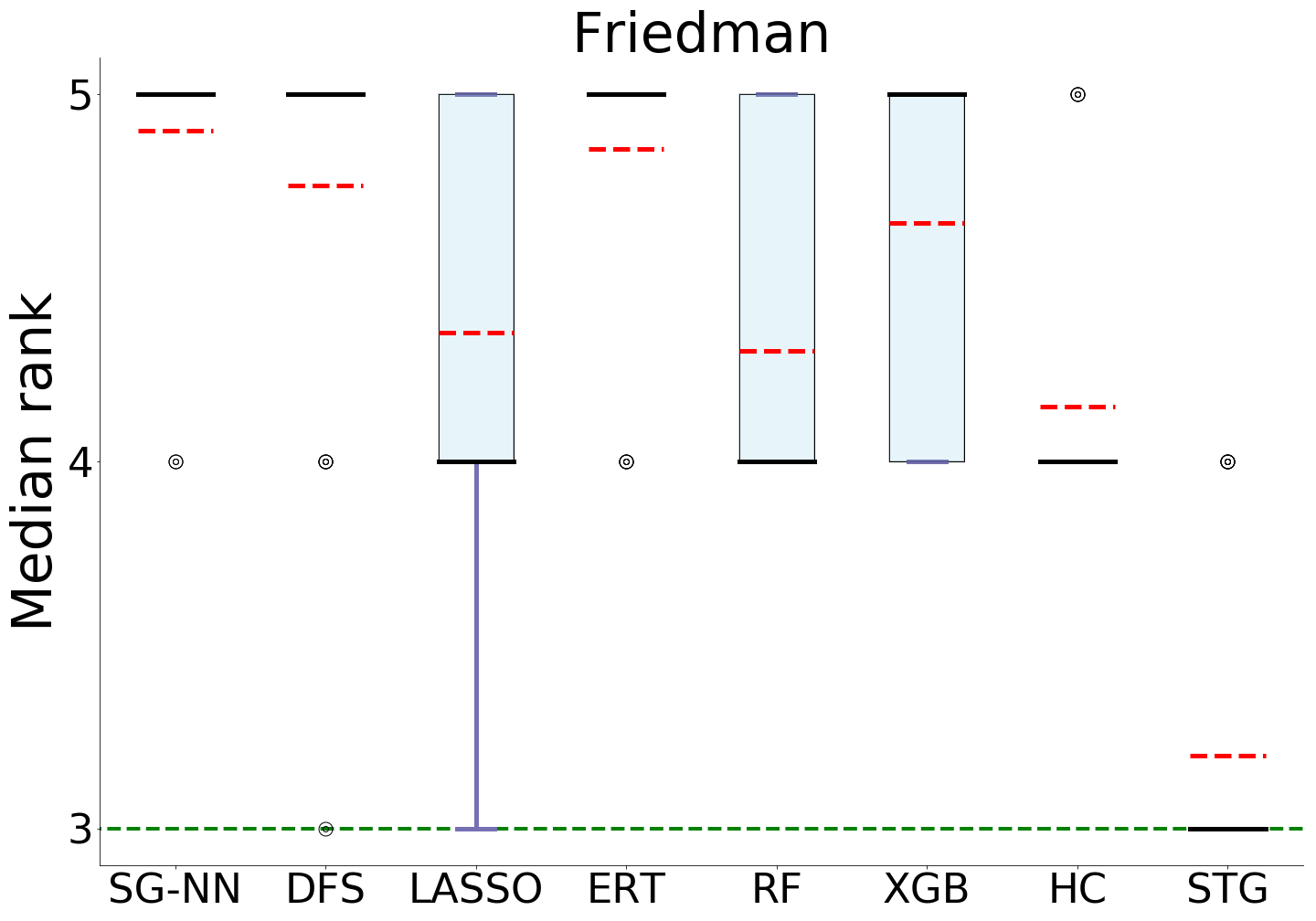}
\includegraphics[width=0.32\textwidth] {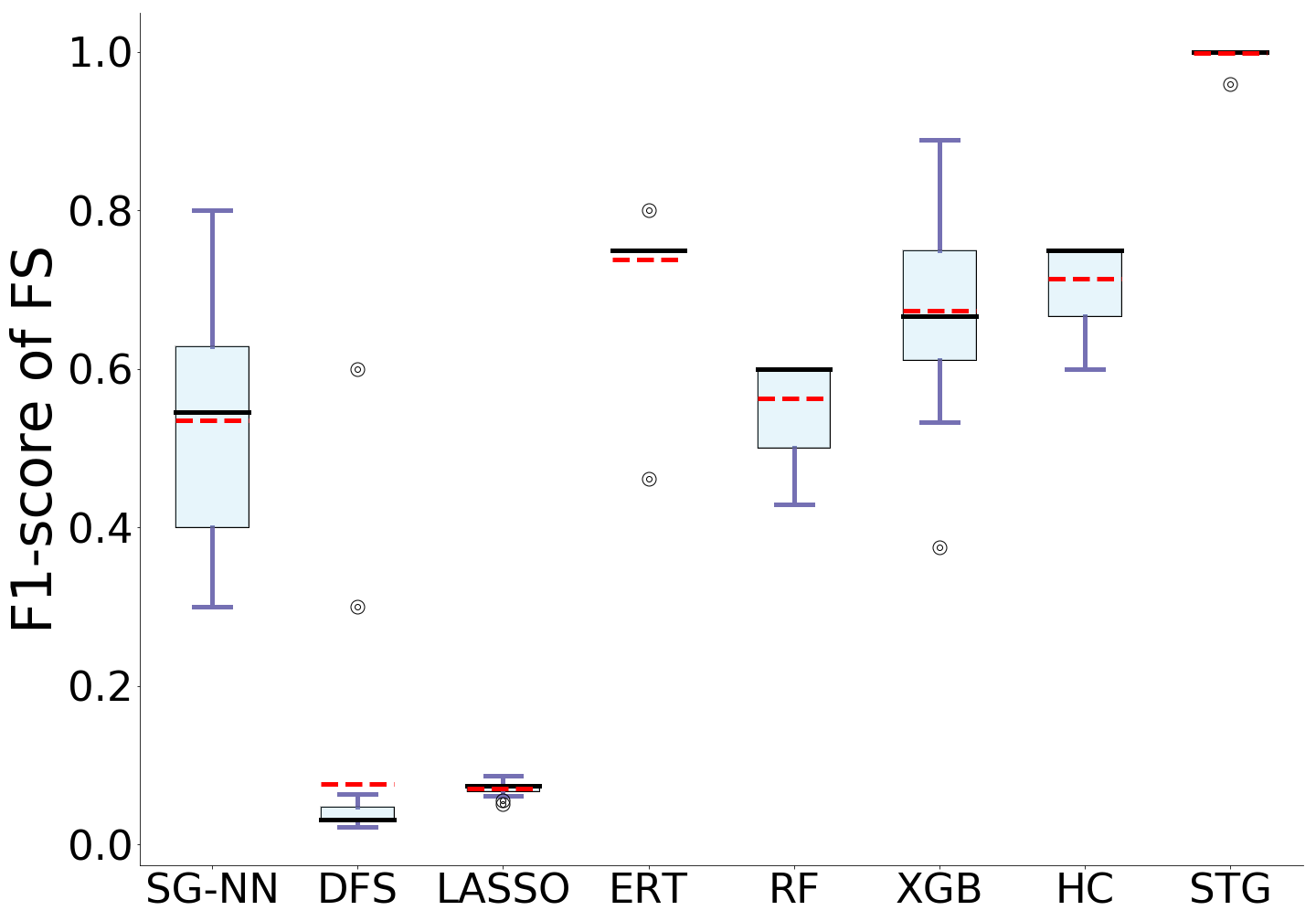}

\end{center}
\vskip -0.2in
\caption{Evaluation of the proposed method using synthetic data. Top row: classification using the MADELON dataset with 5 informative and 495 nuisance features.
Bottom row: regression using a modified version of the Friedman dataset, which also consists of 5 informative and 495 nuisance features.
Left column: Accuracy/root mean squared error (RMSE). Middle column: median rank of informative features. Right column: F1-score that measures success in retrieving the informative features. We also evaluated the accuracy and MSE of a neural network with no feature selection (DNN). Black bars represent the medians, and dashed red lines are the means. In the middle column, dashed green lines are the optimal median ranks.} 
\label{fig:Madelon}
\vskip -0.13in
\end{figure*}

\subsection{Introducing randomness}
Under mild assumptions, we show that one can replace the
deterministic search over the set $\cal{S}$ (or corresponding indicator vector
$\myvec{s}$) with a search over the parameters of the distributions that model
$\myvec{s}$. Our proposition is based on the following two assumptions:
\paragraph{Assumption 1:} There exists a subset of indices $\cal{S}^*$ with a cardinality equal to $k$ such that for any $i \in \cal{S}^*$ we have $I(X_i ; Y | \myvec{X}_{\setminus \{i\}}) > 0$.
\paragraph{Assumption 2:}  $I(\myvec{X}_{{\cal{S}^*}^c} ; Y | \myvec{X}_{\cal{S}^*}) = 0$.
\paragraph{Discussion of assumptions:} Assumption 1 states that including an element from $\cal{S}^*$ improves prediction accuracy. This assumption is equivalent to stating that feature $i$ is strongly relevant~\cite{kohavi1997wrappers,brown2012conditional}. Assumption 2 simply states that $\cal{S}^*$ is a superset of the Markov Blanket of the variable $Y$~\cite{brown2012conditional}. The assumptions are quite benign. For instance, they are satisfied if $\myvec{X}$ is drawn from a Gaussian with a non-degenerate covariance matrix and $Y = f(\myvec{X}_{\cal{S}^*}) + w$, where $w$ is noise independent of $\myvec{X}$ and $f$ is not degenerate. With these assumptions in hand, we may present our result.

\begin{props}\label{prop:1}
Suppose that the above assumptions hold for the model. Then, solving the optimization~\eqref{eq:MI_c} is equivalent to solving the optimization
\begin{equation}\label{eq:beropt}
\max_{\myvec{0} \leq \myvec{\pi} \leq \myvec{1}} I(\myvec{X} \odot {\myvec{\tilde{S}}}; \myvec{Y})   \quad  \text{s.t.} \quad \sum_i \exval[\tilde{S}_i] \le k,
\end{equation}
where the coordinates $\tilde{S}_i$ are drawn independently at random from a Bernoulli distribution with parameter $\pi_i$.
\end{props} Due to length constraints, we leave the proof of this proposition and how it bridges the MI maximization~\eqref{eq:MI_c} and risk minimization~\eqref{eq:1} \rev{to} the supplementary material.   

\section{Related Work}

The three most related works to this study are \cite{Louizos2017LearningSN}, \cite{jordan} and \cite{Chang}. 
In \cite{Louizos2017LearningSN}, they introduce the Hard-Concrete (HC) distribution as a continuous surrogate for Bernoulli distributions in the context of model compression.  
The authors demonstrate that applying the HC to all of the weights leads to fast convergence and improved generalization. They did not evaluate the HC for the task of feature selection where stability of the selection is an important property.

In \cite{jordan}, the Concrete distribution is used to develop a framework for interpreting pre-trained models. 
Their method is focused on finding a subset of features given a particular sample and, therefore, is not appropriate for general feature selection. In \cite{Chang}, the Concrete distribution is used for feature ranking. The method is not fully embedded and requires model retraining to achieve feature selection.

Bernoulli relaxation techniques that are based on logistic distributions (e.g. Concrete/Gumbel-Softmax and HC) are not suitable for feature selection. Specifically, the use of the Concrete/Gumbel-Softmax distribution ranks features but retains all of them (no feature selection). 
In contrast to our Gaussian-based relaxation of Bernoulli distributions (STG),
the logistic-based HC yields high-variance gradient estimates. 
For model sparsification, this high variance is not problematic because the sparsity pattern within the network does not matter as long as the method achieves enough sparsity as a whole. 
For feature selection based on the HC approach, however, the subsets of selected features at different runs vary substantially. Thus, the stability of the HC-based feature selection is poor; see Section \ref{sec:discussion}.
Furthermore, higher gradient variance will also result in a slower SGD convergence, which has been demonstrated empirically in Section \ref{sec:experiments} and the supplementary material.

\section{Experiments}
\label{sec:experiments}

\begin{figure*}[htb!]
\vskip -0.1in
\begin{center}
\includegraphics[width=0.32\textwidth] {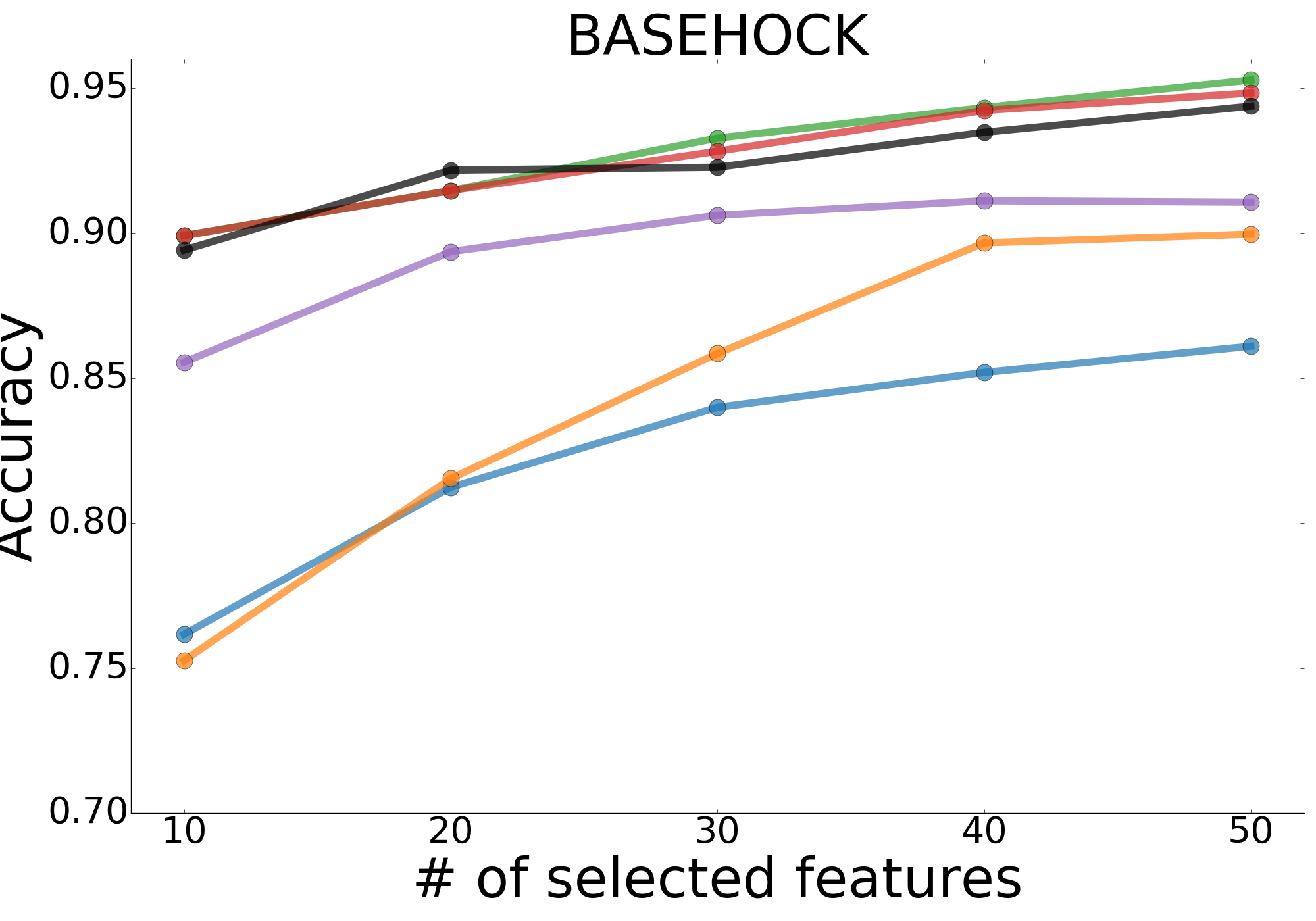}
\includegraphics[width=0.32\textwidth] {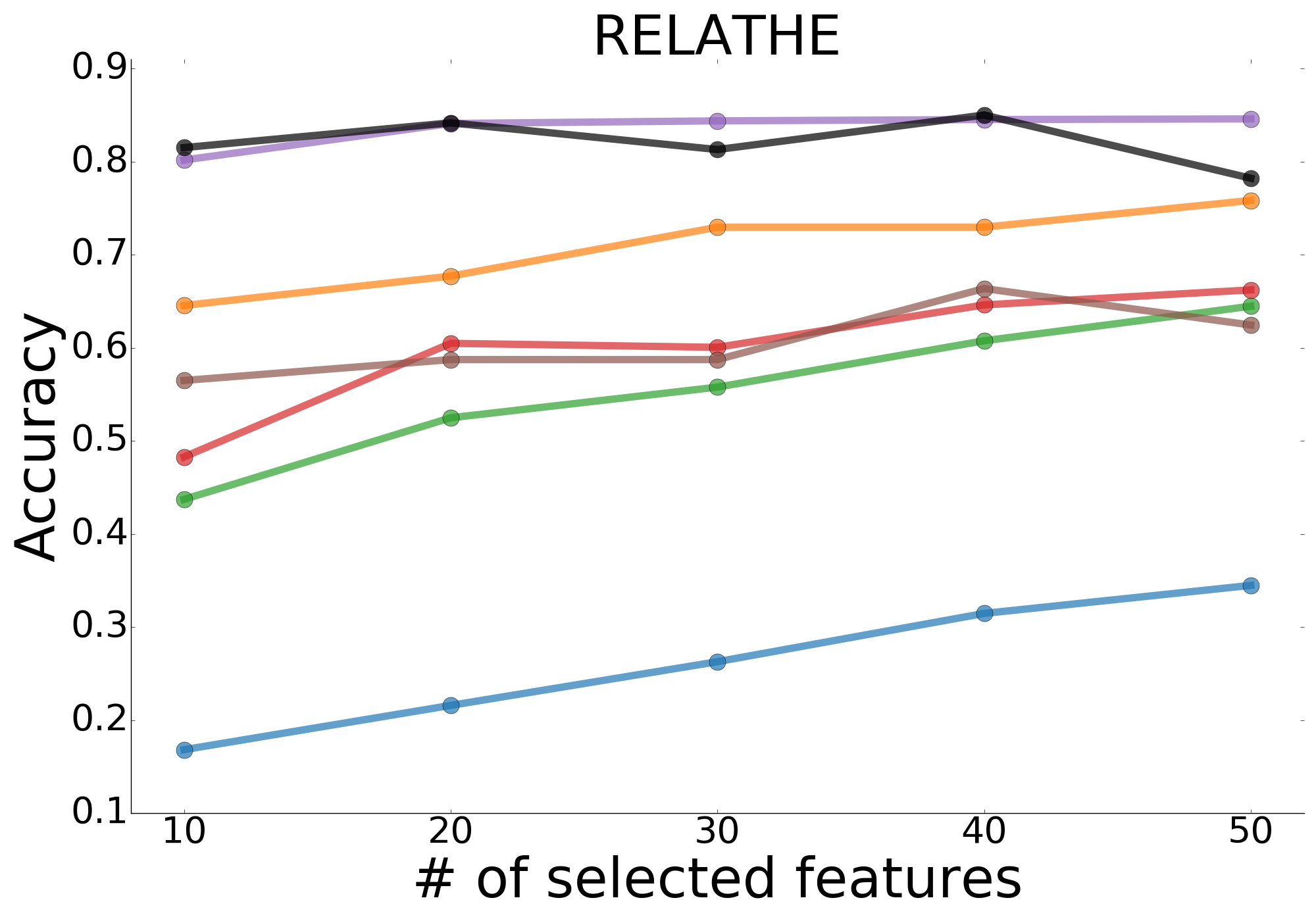}
\includegraphics[width=0.32\textwidth] {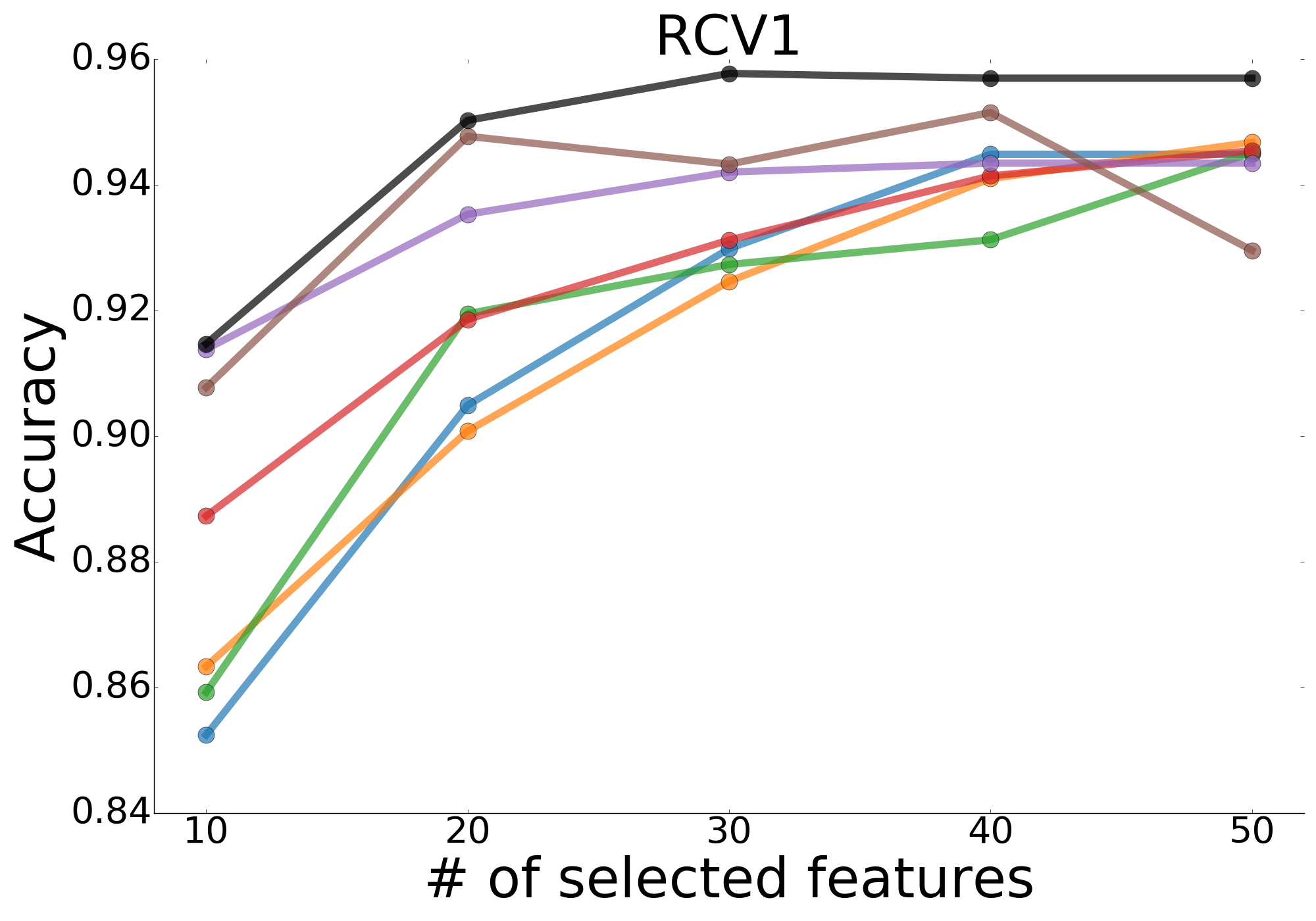}
\includegraphics[width=0.32\textwidth] {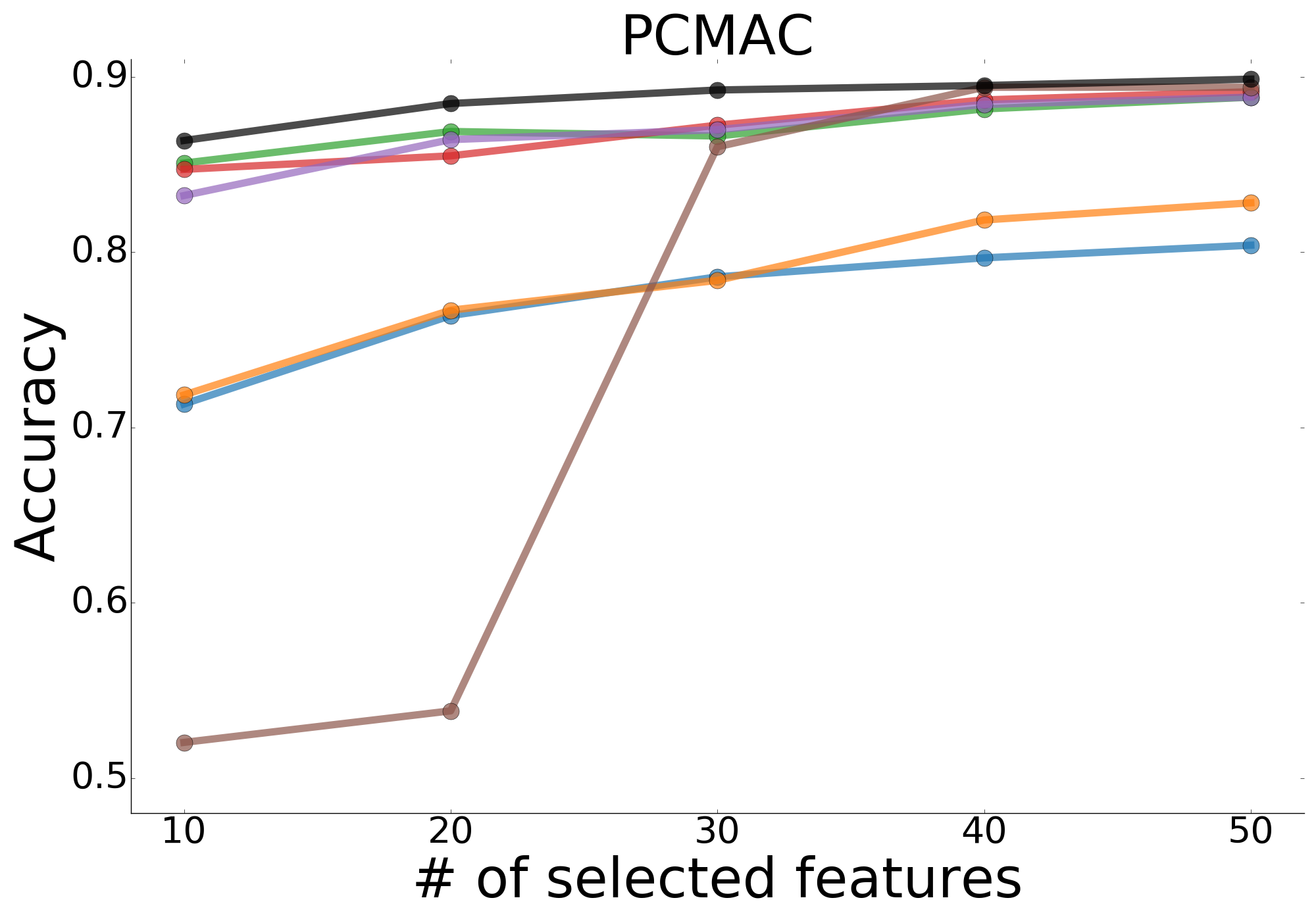}
\includegraphics[width=0.32\textwidth] {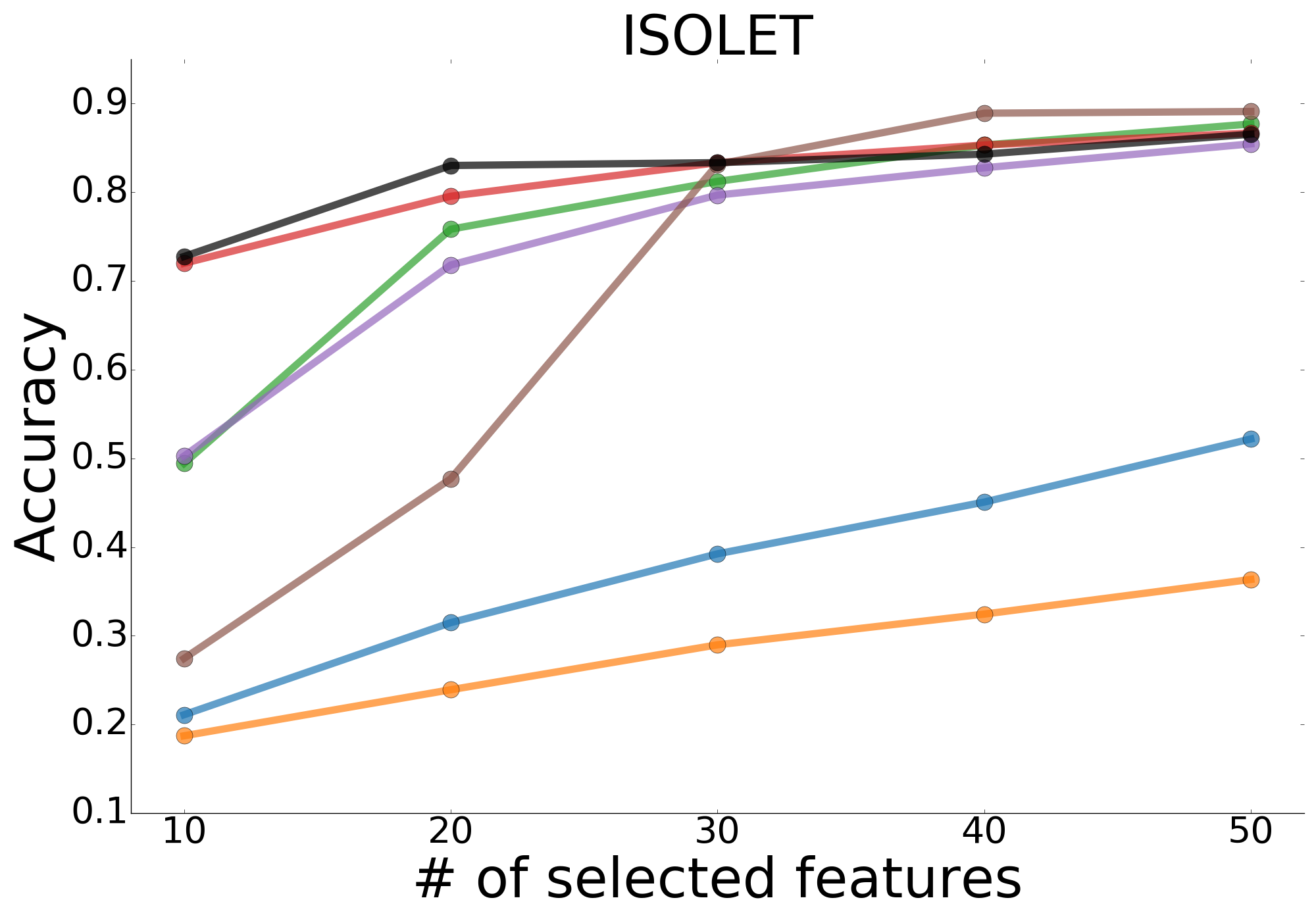}
\includegraphics[width=0.32\textwidth] {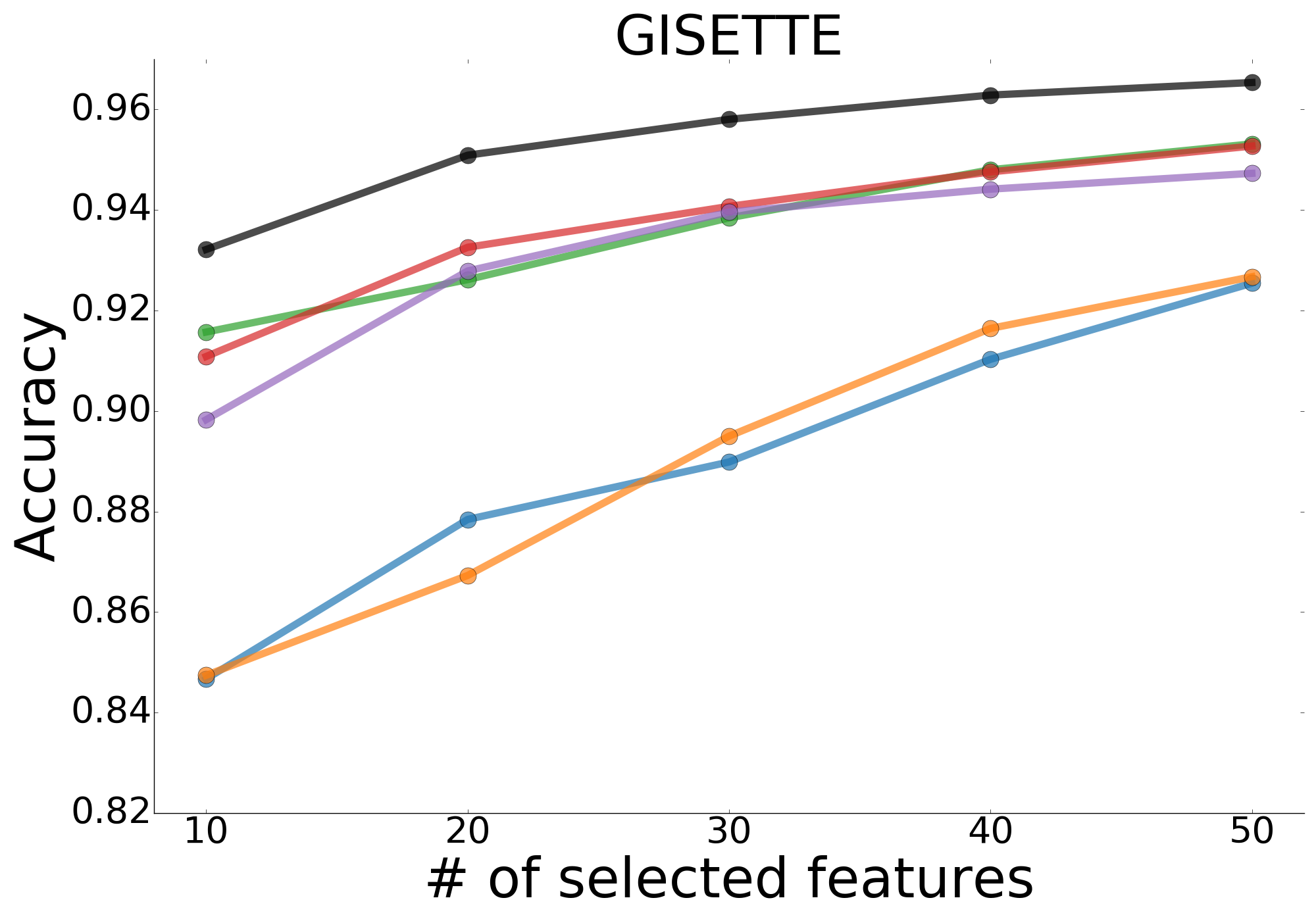}
\includegraphics[width=0.32\textwidth] {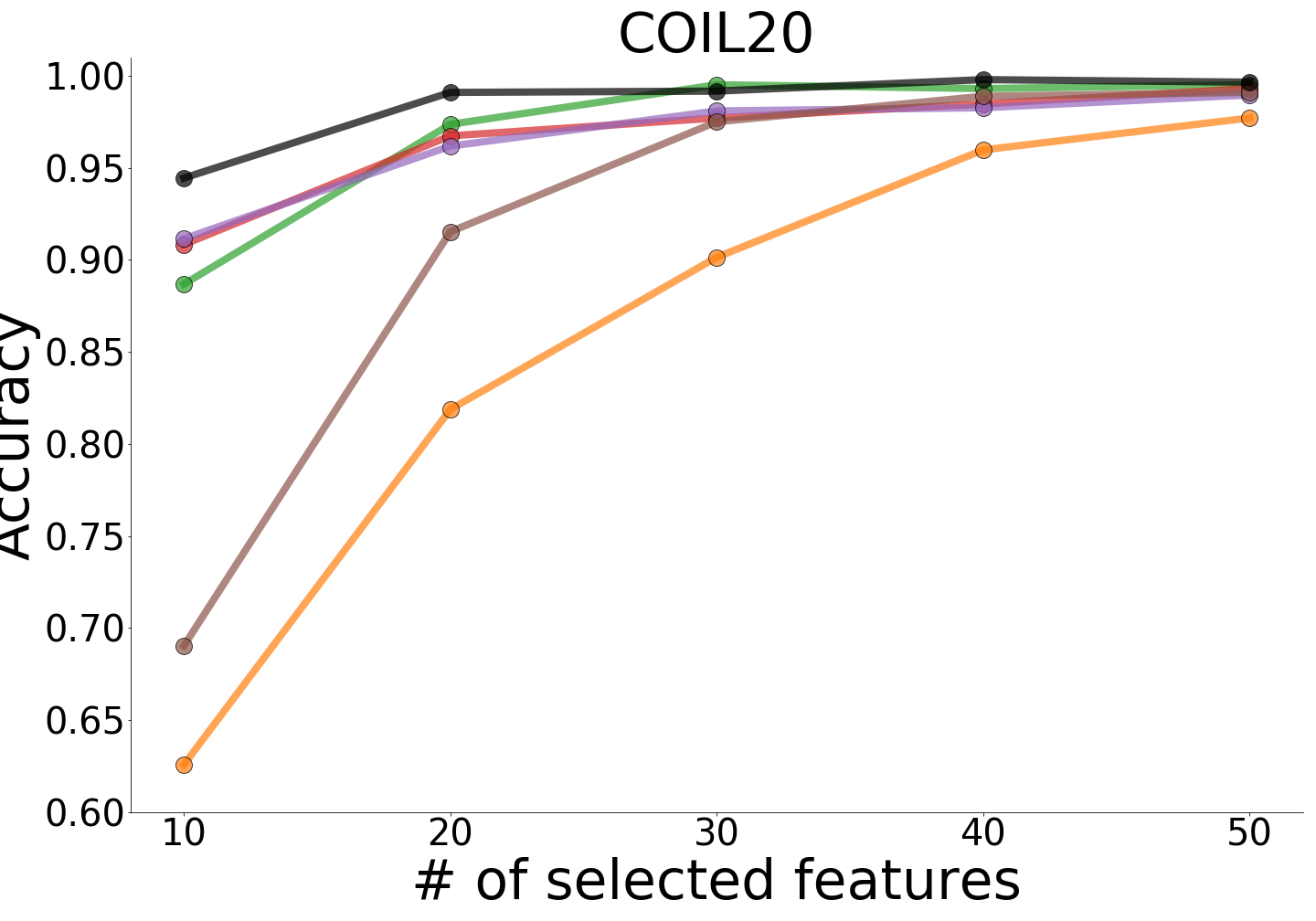}
\includegraphics[width=0.32\textwidth] {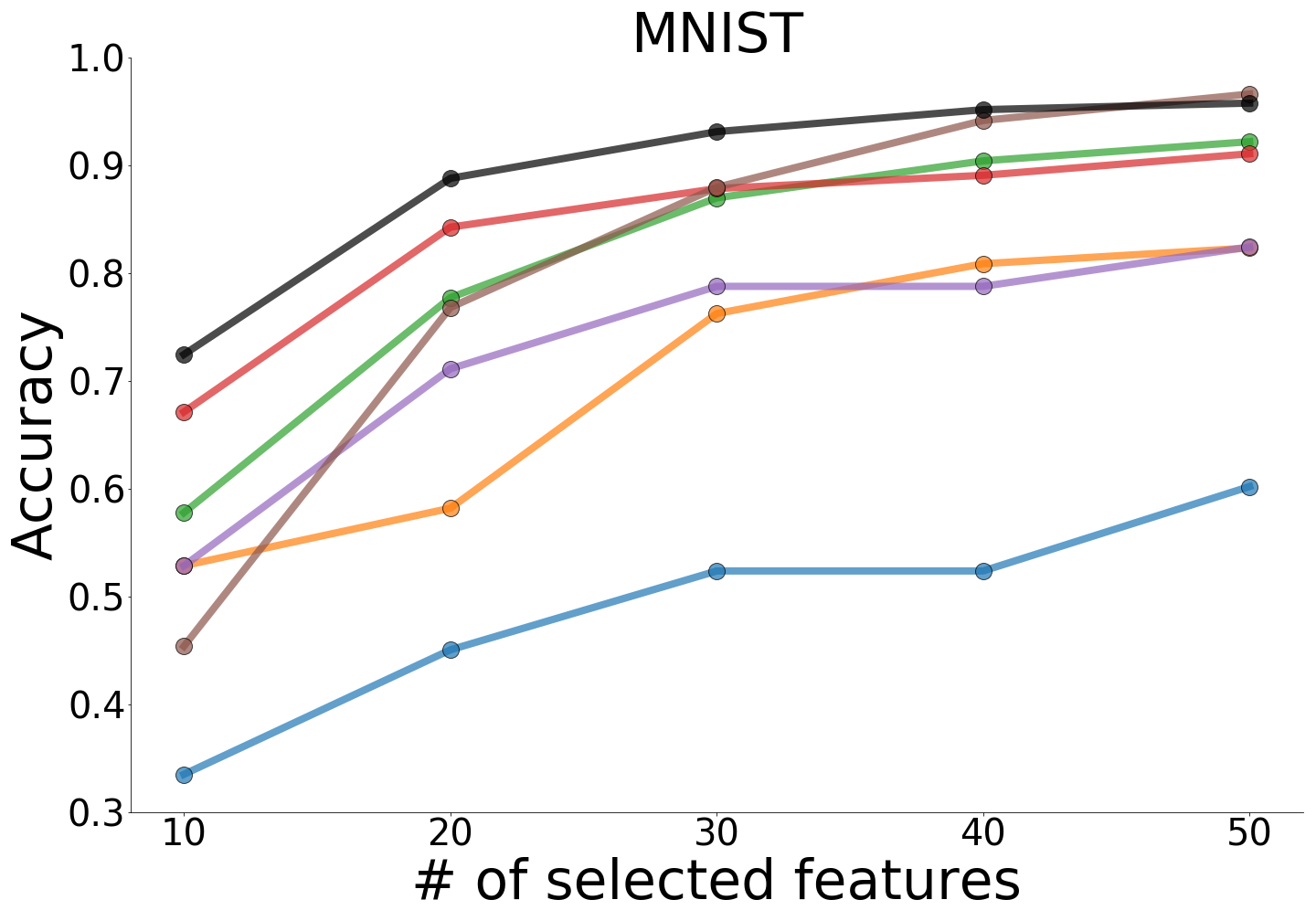}
\includegraphics[width=0.32\textwidth]{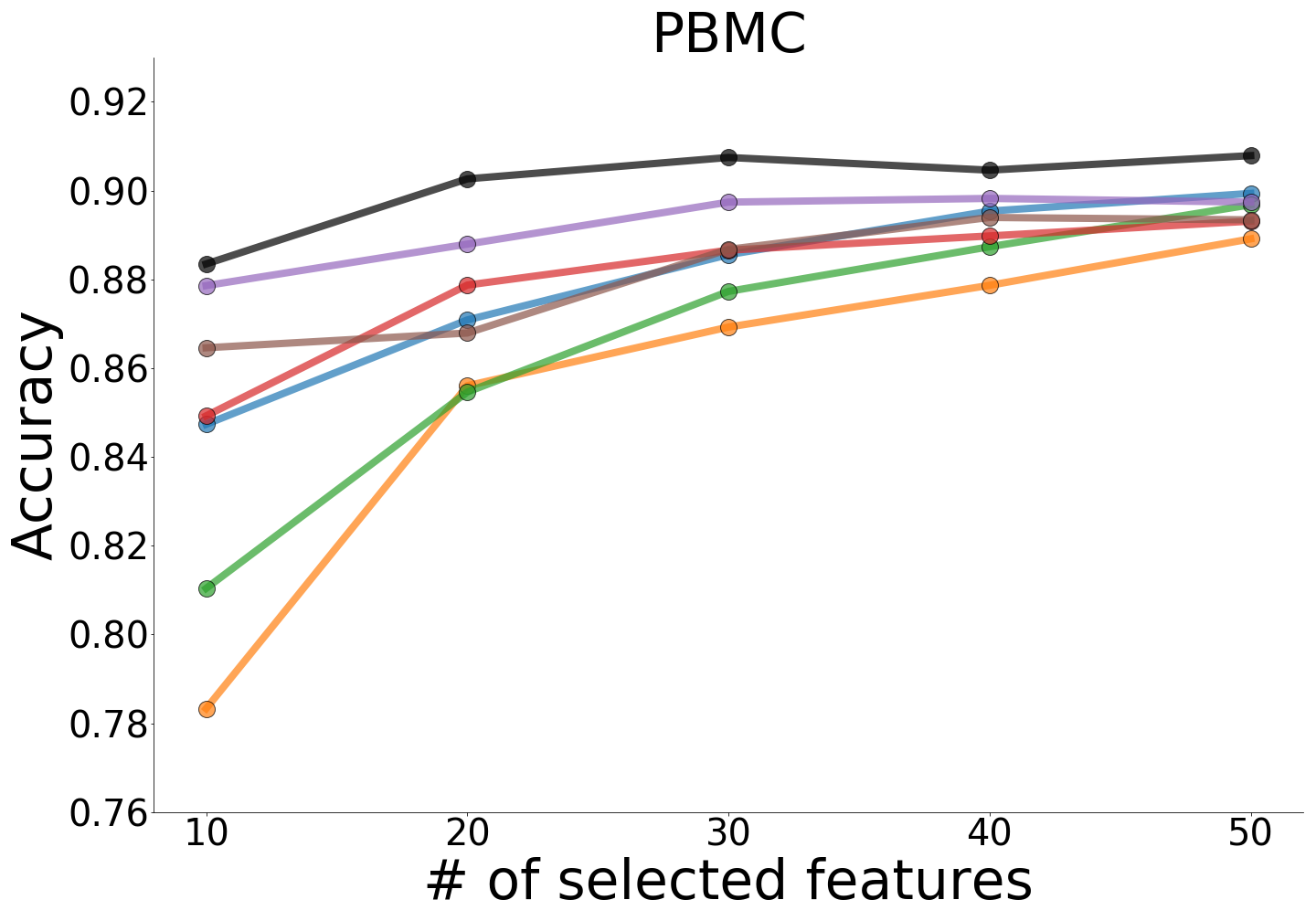}
\includegraphics[width=1\textwidth]{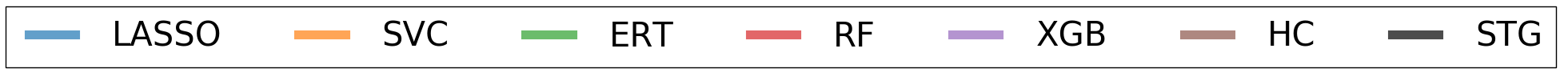}
\end{center}
\vskip -0.2in
\caption{ Classification accuracy vs. number of selected features. Descriptions of the 9 datasets appear in Table \ref{tab:realdata}. } 
\label{fig:realdata}
\vskip -0.08in
\end{figure*}
Here, we evaluate our proposed embedded feature selection method. We implemented \footnote{https://github.com/runopti/stg} it using both the STG and HC \cite{Louizos2017LearningSN} distributions and tested on several artificial and real datasets. We compare our method with several classification and regression algorithms including embedded methods such as LASSO \cite{Lasso}, linear support vector classification (SVC) \cite{SVC}, deep feature selection (DFS) \cite{DFS} and group-sparse regularization for
deep neural networks (SG-NN) \cite{sparseNN-group-lasso}. Our method is also compared with leading tree-based wrapper methods - extremely randomized trees (ERT) \cite{Tree}, random forests (RF) \cite{Rf1} and XGBoost (XGB) \cite{xgboost}. See the supplementary material for details on the hyper-parameters of all methods.

\subsection{Synthetic data in the $D>N$ regime  }
\label{sec:synth}
Now we present empirical results in the challenging regime where the number of features exceeds the number of samples ($D>N$). We use synthetic datasets with informative and irrelevant nuisance variables. We begin with the MADELON dataset, a hard classification problem suggested in the NIPS 2003 feature selection challenge \cite{nips2003}. This dataset consists of $5$ informative features and $495$ nuisance features. See the supplementary material for additional details on the MADELON dataset.

Next, we present a regression scenario based on a modification of the Friedman regression dataset \cite{friedman}. In this dataset, all $500$ variables are uniformly distributed in $[0,1]$, and the response is defined by the following function: $$ Y=10\sin(X_1 X_2)^2+20X_3^2+10 \text{ sign}(X_4 X_5-0.2)+\xi,$$ where $\xi$ is drawn from $\mathcal{N}(0,1)$. Then, $Y$ is centered and divided by its maximal value.

For the above synthetic classification and regression datasets, we generate $600$ samples of which we use $450$ for training, $50$ for validation and $100$ for a test set. The hyper-parameter (controlling the number of selected features) for each method is optimized based on the validation performance. 
The experiment is repeated $20$ times, and the accuracy/root mean squared error (RMSE), median rank and F1-score that measures feature selection performance are presented in Fig. \ref{fig:Madelon}. 
To compute the median ranks, we utilize the scores that each method assigns to the features. We then rank all features based on these scores and compute the median of the ranks of the $5$ informative features.
Thus, the optimal median rank in these examples is $3$. The F1-score measure for feature selection is defined as $F1=2{(\text{precision} \cdot \text{recall}) }/{(\text{precision} + \text{recall})}$, where precision and recall are computed by comparing the selected/removed features to the informative/nuisance features. For example, a model which retains all of the  features has a recall of $1$ with a precision of ${5}/{500}$.

The results presented in Fig. \ref{fig:Madelon} demonstrate our embedded method's ability to learn a powerful predictive model in the regime of $D>N$, where the majority of variables are not informative. Even though our median rank performance is comparable to tree-based methods, we outperform all baseline in F1-scores. 
%In terms of feature selection capabilities, even though our performance is comparable to tree based methods in terms of the median rank, we outperform all alternative methods in terms of the F1-score of feature selection. 
This demonstrates that our embedded method is a strong candidate for the task of finding complex relations in high dimensional data.

\begin{table*}[htbp!]
\vskip -0.2in
  \centering
  \caption{Description of the real-world data used for empirical evaluation.}
  \vskip -0.01in
   \begin{adjustbox}{width=\textwidth,center}
    \begin{tabular}{lrrrrrrrrr}
    \toprule
          & \multicolumn{1}{l}{BASEHOCK} & \multicolumn{1}{l}{RELATHE} & \multicolumn{1}{l}{RCV1} & \multicolumn{1}{l}{PCMAC} & \multicolumn{1}{l}{ISOLET} & \multicolumn{1}{l}{GISETTE} & \multicolumn{1}{l}{COIL20} & \multicolumn{1}{l}{MNIST} & \multicolumn{1}{l}{PBMC} \\
    \midrule
    Features (D) & 7862  & 4322  & 47236 & 3289  & 617   & 5000  & 1024  & 784   & 17126 \\
    Train size & 1594  & 1141  & 2320  & 1554  & 1248  & 5600  & 1152  & 60000 & 2074 \\
    Test size & 398   & 285   & 20882 & 388   & 312   & 1400  & 288   & 10000 & 18666 \\
    Classes & 2     & 2     & 2     & 2     & 26    & 2     & 20    & 10    & 2 \\
    Data type & Text  & Text  & Text  & Text  & Audio & Image & Image & Image & scRNA-seq \\
    \bottomrule
    \end{tabular}%
    \end{adjustbox}
  \label{tab:realdata}%
  \vskip -0.1in
\end{table*}%
We encourage the reader to look at the supplementary material, where we provide additional experiments including a more challenging variant of the MADELON data, the XOR data, the two-moons data and $3$ artificial regression datasets based on \cite{SRFF}.

\subsection{Noisy Binary XOR Classification}

\label{sec:xor}

In the following evaluation, we consider the problem of learning a binary XOR function for classification task. 
The first two coordinates $x_1, x_2$ are drawn from a "fair" Bernoulli distribution. The response variable is set as an XOR of the first coordinates, such that $y=x_1 \oplus x_2 $. 
The coordinates $x_i,i=3,...,D$ are nuisance features, also drawn from a binary "fair" Bernoulli distribution. 
The number of points we generate is $N=1,500$, of which  $70~\%$ are reserved for test and $10 \%$ of the remaining training set was reserved for validation. We compare the proposed method to four embedded feature selection methods (LASSO \cite{Lasso}, C-support vectors (SVC) \cite{c-sopport}, deep feature selection (DFS) \cite{DFS}, sparse group regularized NN (SG-L1-NN) \cite{sparseNN-group-lasso}).
To provide more benchmarks, we also compare our embedded method against three wrapper methods (Extremely Randomized Trees (Tree) \cite{Tree}, Random Forests (RF) \cite{Rf_use}) and Extreme Gradient Boosting (XGBOOST) \cite{xgboost}.

To evaluate the feature selection performance, we calculate the Informative Features Weight Ratio (IFWR). IFWR is defined as the sum of weights $W_d$ over the informative features divided by the sum over all weights. In the case of binary weights the IFWR is in fact a recall measure for the relevant features.

\begin{figure}[htb!]
\begin{center}
\vskip -0in
\subfigure[]{\label{fig:xor1}\includegraphics[width=0.45\textwidth] {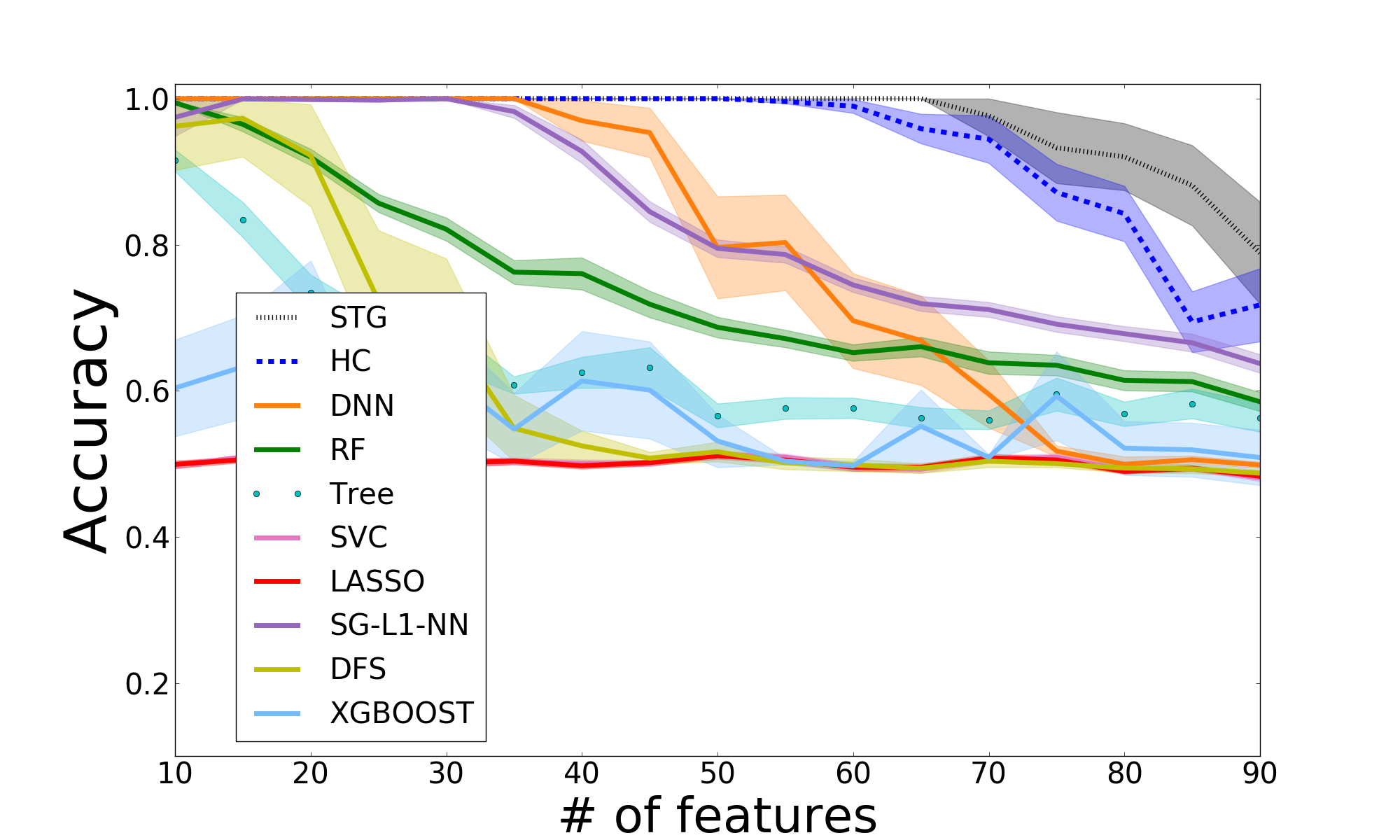}}
\subfigure[]{\label{fig:xor2}\includegraphics[width=0.45\textwidth] {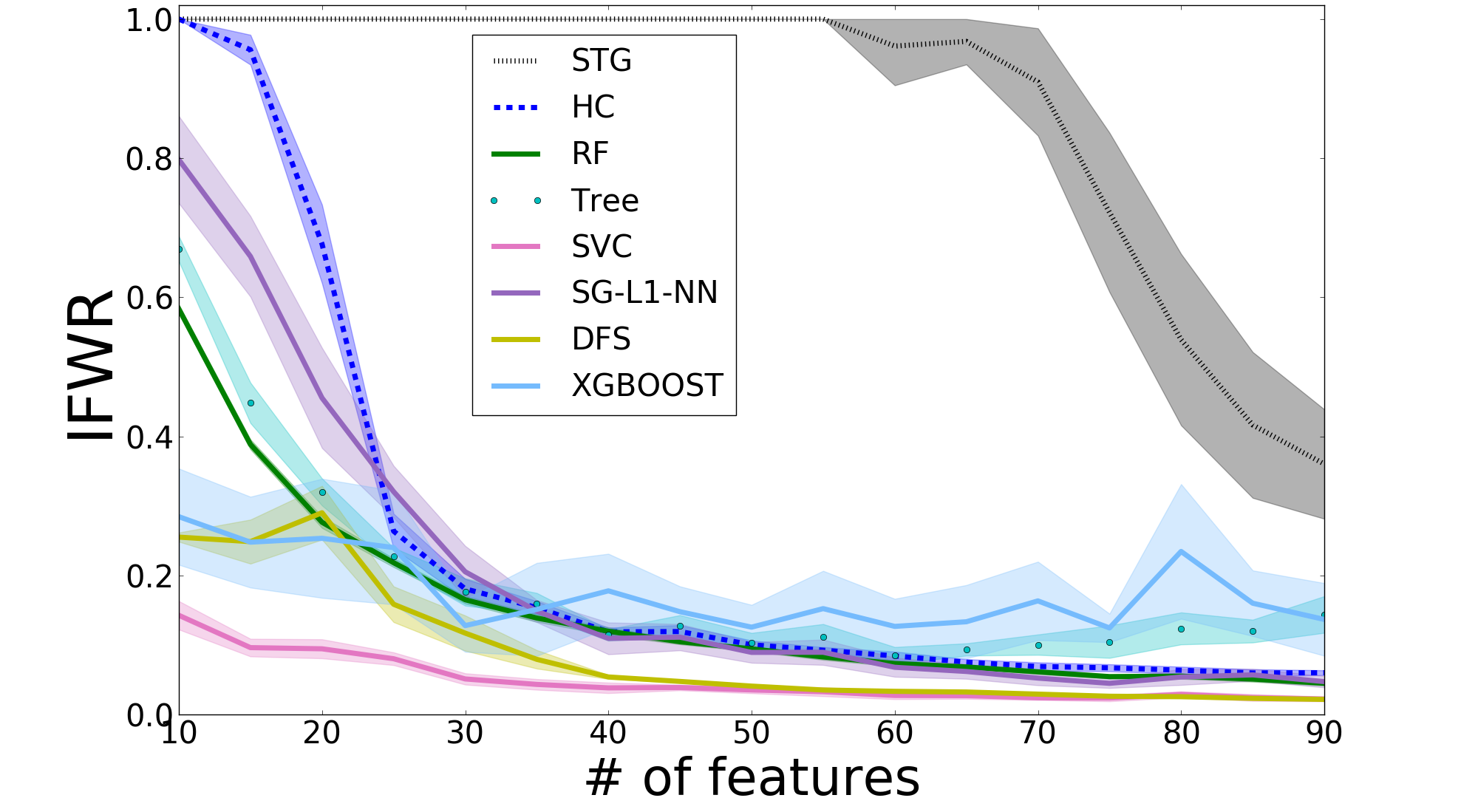}}
\subfigure[]{\label{fig:xor3}\includegraphics[width=0.45\textwidth] {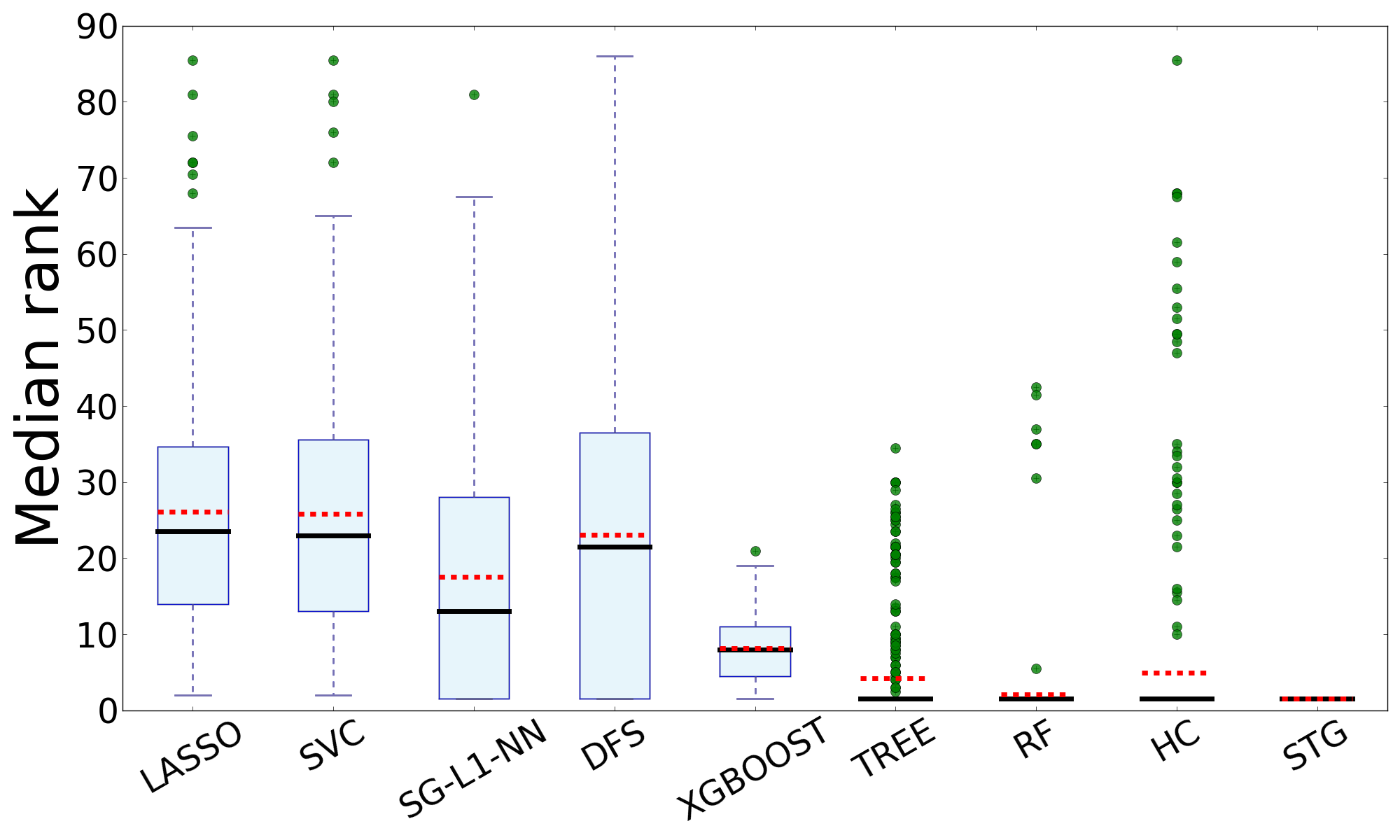}}
\end{center}
\vskip -0in
\caption{\subref{fig:xor1} Classification accuracy (mean and standard deviation) vs. the number of irrelevant noisy dimension ($D$) for the XOR problem. \subref{fig:xor2} The Informative Features Weight Ratio (IFWR), central lines are the means while shaded are represent the standard deviation. IFWR is the sum of weights $W_d$ over the informative features divided by the sum over all weights. \subref{fig:xor3} Box plots for the median rank of the two informative features. Black line and dashed red represent the median and mean of each method. Optimal median rank in this experiment is $1.5$.}
\vskip -0in
\end{figure}

The experiment is repeated $20$ times for different values of $D$, and the average test classification accuracy and standard deviation are presented in Fig. \ref{fig:xor1}, followed by the IFWR in Fig. \ref{fig:xor2}. The number of selected features affects the accuracy. Therefore, to treat all the methods in a fair manner, we tune the hyperparameter that controls the sparsity level using Optuna \cite{OPTUNA} which optimizes the overall accuracy across different $D$s. For instance, the wrapper methods (Tree, RF and XGBOOST) has a threshold value to retain features. We retrain them using only such features whose weight is higher than the threshold. In terms of feature ranking (see Fig. \ref{fig:xor3}, only the tree based methods and the proposed (STG and HC based) provide the optimal median rank (which is $1.5$) for the two relevant features. Moreover, the ranking provided by STG is the most stable comparing to all the alternative methods. 

\subsection{Classification on real world data}
We now turn our attention towards using several real-world labeled datasets to evaluate our method. Most of the datasets are collected from the ASU feature selection database available online \footnote{http://featureselection.asu.edu/datasets.php}. The dimensions, sample size and domains of the data are versatile and are detailed in Table \ref{tab:realdata}. On all of the datasets - except MNIST \cite{mnist1}, RCV-1 \cite{rcv1} and the PBMC \cite{pbmcP} - we perform 5-fold cross validation and report the average accuracies vs. the number of features that the model uses for several baselines. 
For MNIST, RCV-1 and the PBMC - we used prefixed training and testing sets which are described in the supplementary material. The results are presented in Fig. \ref{fig:realdata}.

Compared to the alternative linear embedded methods (i.e. LASSO and SVC), the nonlinearity of our method provides a clear advantage. 
While there are regimes in which the tree-based methods slightly outperform our method, they require retraining a model based on the selected features; however, our method is only trained once and learns the model and features simultaneously. 
This experiment also demonstrates that the STG is more suited for the task of feature selection than the HC. Note that in this experiment we did not include the DFS and SG-NN, as they do not sparsify the weights and, therefore, cannot be evaluated vs. the number of selected features. 

Due to lack of space we leave the real word regression experiments to the supplementary material.

\section{Cox Proportional Hazard Models for Survival Analysis}  \label{sec:cox}

A standard model for survival analysis is the Cox Proportional Hazard Model. In \cite{Jared}, the authors proposed DeepSurv that extends the Cox model to neural networks. 
We incorporate our method into DeepSurv to see how our procedure improves survival analysis based on gene expression profiles from the breast cancer dataset called METABRIC \cite{metabric} (along with additional commonly used clinical variables.) 
See the supplementary material for more details about the dataset and experimental setup.

We compare our method Cox-STG with four other methods: a Cox model with $\ell_1$ regularization (Cox-LASSO), Random Survival Forest (RSF) \cite{RSF}, Cox-HC, and the original DeepSurv. We evaluate the predictive ability of the learned models based on the concordance index (CI), a standard performance metric for model assessment in survival analysis; it measures the agreement between the rankings of the predicted and observed survival times. The performance of each model in terms of the CI and the number of selected features are reported in Table \ref{tab:cox}.
The Cox-STG method outperforms the other baselines indicating that our approach identifies a small number of informative variables while maintaining high predictive performance. 

\begin{table*}[htb!]
\vskip -0.1in
\caption{Performance comparison of survival analysis on METABRIC. We repeat the experiment 5 times with different training/testing splits and report the mean and standard deviation on the testing set.}
\label{tab:cox}

\begin{center}
\vskip -0.08in
\begin{small}
\begin{sc}
\begin{tabular}{lcccccc}
\toprule
 & DeepSurv& RSF& Cox-Lasso  & Cox-HC & Cox-STG \\
\midrule
C-index &0.612 (0.009) & 0.626 (0.006) &  0.580 (0.003) & 0.615 (0.007)& \textbf{0.633} (0.005)  \\
$\#$ features &  221 (All) & 221 (All) & 44 (0) & 14 (1.72) & 2 (0) \\
\bottomrule
\end{tabular}
\end{sc}
\end{small}
\end{center}
\vskip -0.12in
\end{table*}

\section{Evaluating stochastic regularization schemes}
\label{sec:discussion}

In this section, we elaborate on two aspects of our proposed method that lead to performance gains: (i) benefits of our non-convex regularization and injected noise, and (ii) advantages of the Gaussian based STG over the logistic based HC distribution in terms of feature selection performance.

To demonstrate these performance gains, we perform a controlled experiment in a linear regression setting. We first generate the data matrix, $\myvec{X} \in \mathbb{R}^{N \times D}$, $D=64$, with values randomly drawn from $\mathcal{N}(0,1)$ and construct the response variable \begin{equation}
\myvec{y} = \myvec{X} \myvec{\beta}^* + \myvec{w},\end{equation} where the values of the noise $\myvec{w}_i,i=1,...,N$ are drawn independently from $\mathcal{N}(0,0.5)$. As suggested by \cite{wainwright09}, we use a known sparsity $\norm{\myvec{\beta}^*}_0 = k$, set by $k=\lceil 0.4 D^{0.75} \rceil=10$. For each number of samples $N$ in the range $[10,250]$, we run $200$ simulations and count the portion of correctly recovered informative features (i.e. the support of $\myvec{\beta}^*)$. For LASSO, the regularization parameter was set to its optimal value $\alpha_N=\sqrt{\frac{2\sigma^2 \log(D-k)\log(k)}{N}}$ \cite{wainwright09}. For STG and HC, we set $\lambda_N=C \alpha_N$, such that $C$ is a constant selected using a cross validated grid search in the range $[0.1,10]$. To evaluate the effect of non-convex regularization and noise injection, we compare the STG to a deterministic non-convex (DNC) counterpart of our method (see definition below) and LASSO, which is convex. To gain insights on (ii), we also compare the HC.

We define the deterministic non-convex (DNC) objective as
\begin{equation}
\label{eq:det}
   \min_{\myvec{\theta},\myvec{\mu}} \frac{1}{N} \sum_{n=1}^N  (\myvec{\theta}^T\myvec{x}_n\odot \myvec{\tilde{z}} - y_n)^2 + \lambda \sum_{d=1}^D \Phi \left ( \frac{\mu_d}{0.5} \right),
\end{equation} 
where $\myvec{\Phi}$ is the standard Gaussian CDF. Combined with $\tilde{z}_d$, this non-convex regularized objective is deterministic and differentiable, and its solution can be searched via gradient descent.

\begin{figure}[htb!]
\vskip -0.15in
\begin{center}
\includegraphics[width=0.4\textwidth] {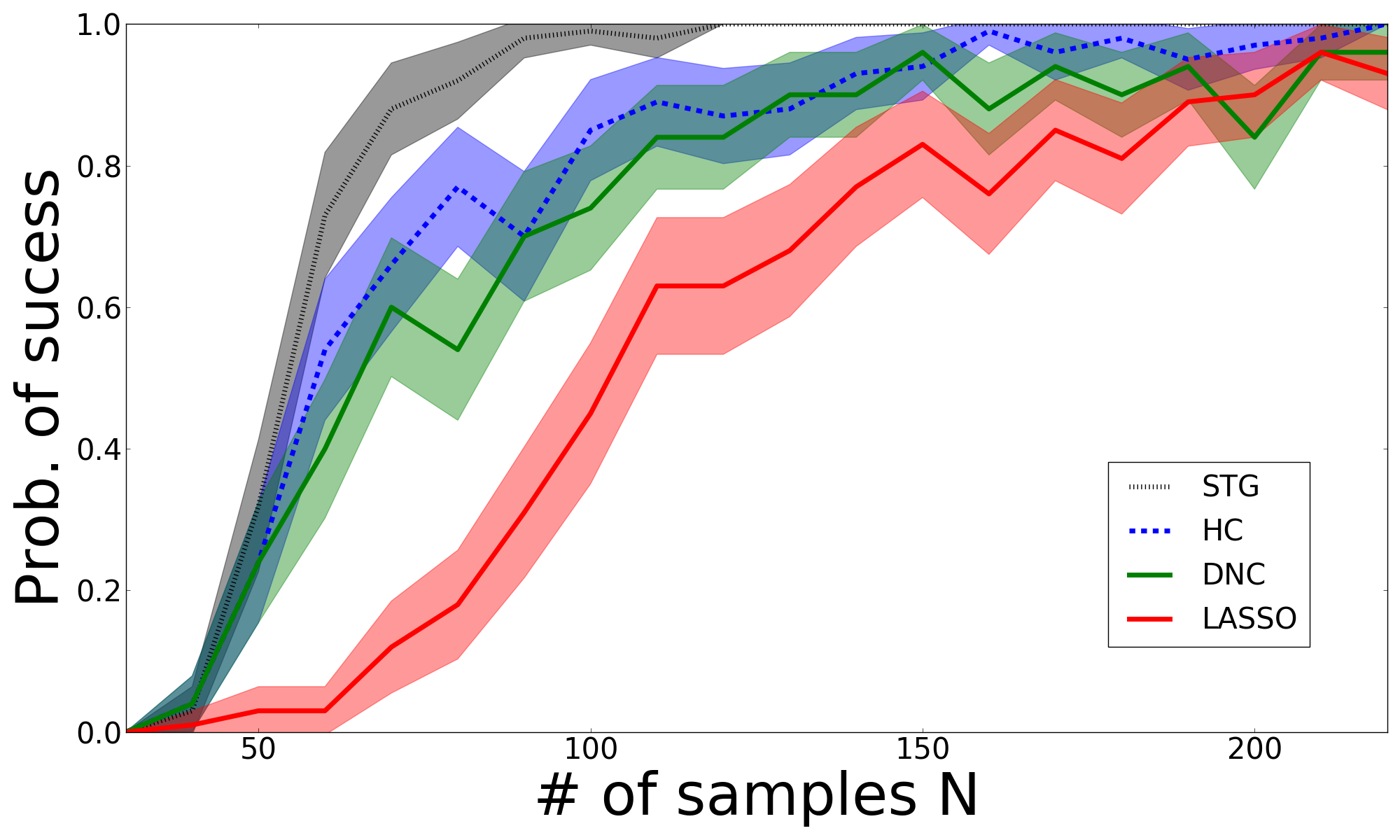}
\includegraphics[width=0.42\textwidth] {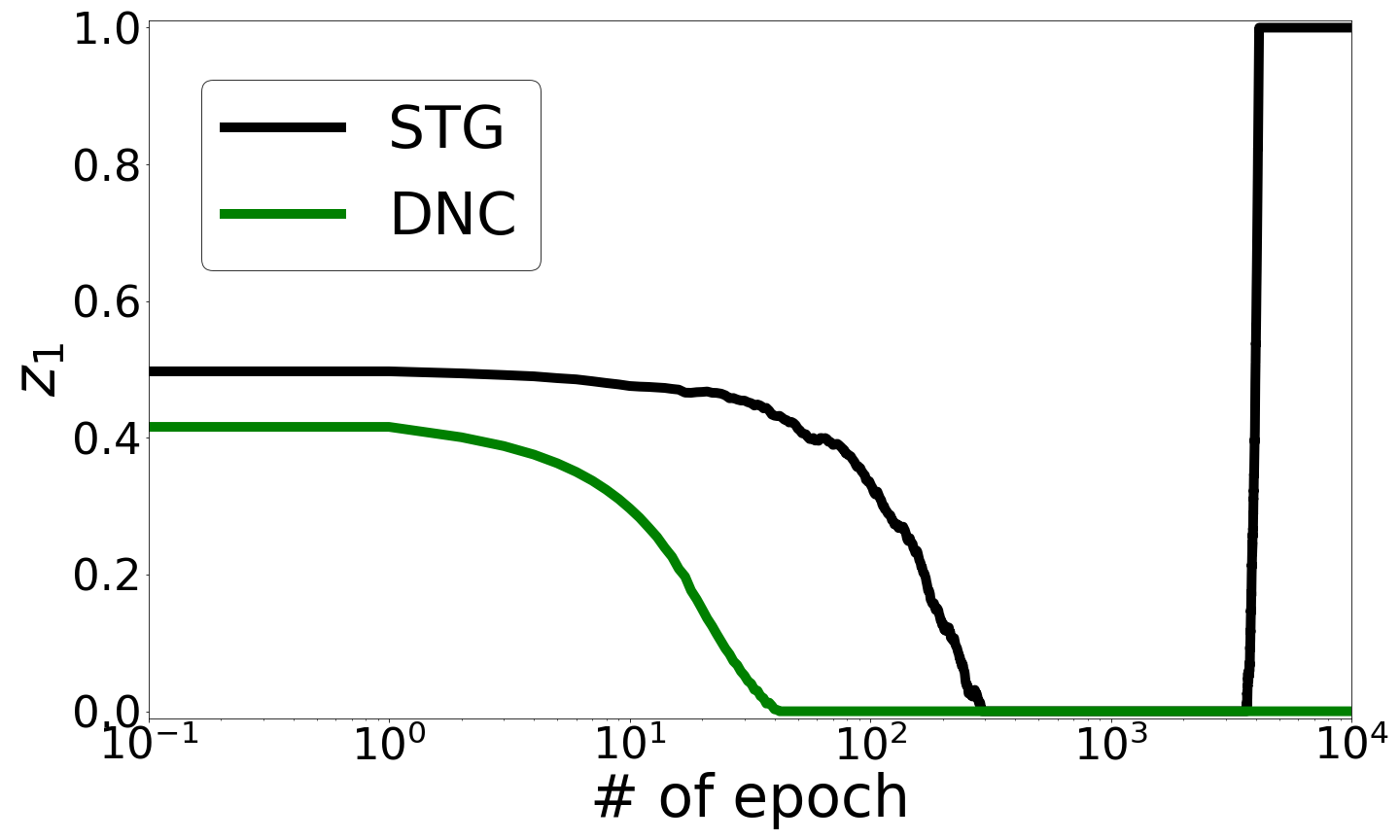}
\end{center}
\vskip -0.2 in
\caption{ Feature selection in linear regression (see Section \ref{sec:discussion}). The goal is to identify the subset of informative features. Top: Probability of recovering the informative features as a function of the number of samples. Comparison between STG, HC, LASSO and DNC. Bottom: The value of a gate $z_1$ throughout training. In STG, injected  noise may lead to a ``second chance'' effect, which in this example occurs after ~$4000$ epochs (black line). In the deterministic DNC setting (green line), a feature's elimination causes its gradient to vanish for the rest of the training. } \vskip 0in
\label{fig:dnc}
\end{figure}

As demonstrated in Fig. \ref{fig:dnc}, the non-convex formulation requires less samples for perfect recovery of informative features than the LASSO. The injected noise based on the HC and STG provides a further improvement. Finally, we observe that the STG is more stable and has a lower variance than the HC, as shown by the shaded colors.

Application of the deterministic formulation is associated with a phenomenon that causes the gradient of an input feature to vanish and \textit{never} acquire a nonzero value if it is zeroed out in an early training phase. In contrast, when we apply STG, a feature that at a certain step has a zero value is not permanently locked because the gate associated with it may change its value from zero to one at a later phase during training. This is due to the injected noise that allows our proposed method to \textit{reevaluate} the gradient of each gate. In Fig. \ref{fig:dnc}, we demonstrate this ``second chance'' effect using $N=60$ samples and presenting the gate's values (throughout training) for an active feature.

\begin{figure}[htb!]
\vskip 0in
\begin{center}

\includegraphics[width=0.4\textwidth] {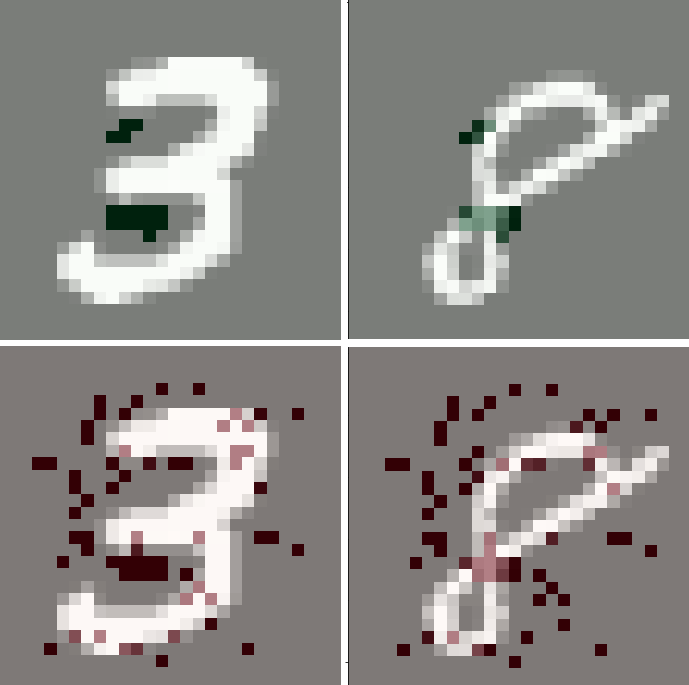}
\end{center}
\vskip 0in
\caption{Comparing stability of feature selection by STG and HC. We train our method to classify $3s$ and $8s$ (from MNIST) with a regularization parameter tuned to retain $\sim 6$ features. We repeat the experiment using $20$ random initializations. Dark pixels represent the union of selected features based on the STG (top) and HC (bottom) overlaid on top of two randomly sampled examples from each class. This demonstrates that an HC-based feature selection does not provide a stable selection of features across different runs. } \vskip 0in
\label{fig:mnist83}
\end{figure}

The advantage of the Gaussian-based STG distribution over the HC distribution stems from the heavier tail of HC, whose form is a logistic distribution. We demonstrate that the heavy-tail distribution is not suitable for feature selection due to its high variance. An ideal feature selection algorithm is expected to identify a consistent set of features across different runs (feature stability), but HC selects many different features in each run resulting in high variance or lack of stability of the selected features.

To further examine the effect of heavy tail distributions, we train two identical neural networks on MNIST but use two different distributions for the gates: Gaussian-STG and HC. Both regularization parameters are tuned to retain $~6$ features. In Fig. \ref{fig:mnist83}, we show that the selected features from the Gaussian-STG are much more consistent across $20$ runs than HC. Furthermore, the variance in the number of selected features is $3.8$ for HC and $1$ for STG. The average accuracies of STG and HC on the test are comparable: $92.4\%$ and $91.7\%$, respectively.

\rev{\section{Feature Selection with Correlations}

Lastly, we evaluate our proposed method using data with correlated features. In real-world, high-dimensional datasets, many features are correlated. Such correlations introduce a challenge for feature selection. For instance, in the most extreme case if there are copies of the same feature, then it is not clear which to select. As another example consider if a large subset of features are a function of a small subset of features that we wish to identify. That large subset of seemingly useful features can confound a feature selection method. Below, we consider a number of examples in various correlated feature settings and demonstrate the strong performance of STG.

%All features are categorized into strongly relevant features, irrelevant features or weakly relevant features. Weakly relevant features are further classified into redundant or non-redundant features. Redundant weakly relevant features may be informative but are correlated with some other informative features. 
%In real-world datasets, the number of redundant weakly relevant features is often much larger than other classes of relevant features. 
%This makes the task of identifying strongly relevant features challenging. Here, we provide an empirical example in which the STG method could be used to identify strongly relevant features in a challenging setting.

We first evaluate the proposed method in a linear setting. 
To introduce correlated features, we extend the linear regression experiment described in Section \ref{sec:discussion} using a correlated design matrix with a covariance matrix whose values are defined by $\Sigma_{i,j} = 0.3^{|i-j|}$.
%for $\rho=0.3$. 
We run $100$ simulations and present the probability of recovering the correct support of $\myvec{\beta}^*$.
Fig. \ref{fig:correlated} shows that even if the features are correlated, STG successfully recovers the support with fewer samples than HC, DNC, and LASSO. 
%XXXXXX add conclusion after we have the plot.

\begin{figure}[htb!]
\vskip -0.16in
\begin{center}
%\subfigure[]{\label{fig:madelon_acc}\includegraphics[width=0.48\textwidth] {Figures/Madelon_new_l_max.png}}
\includegraphics[width=0.48\textwidth] {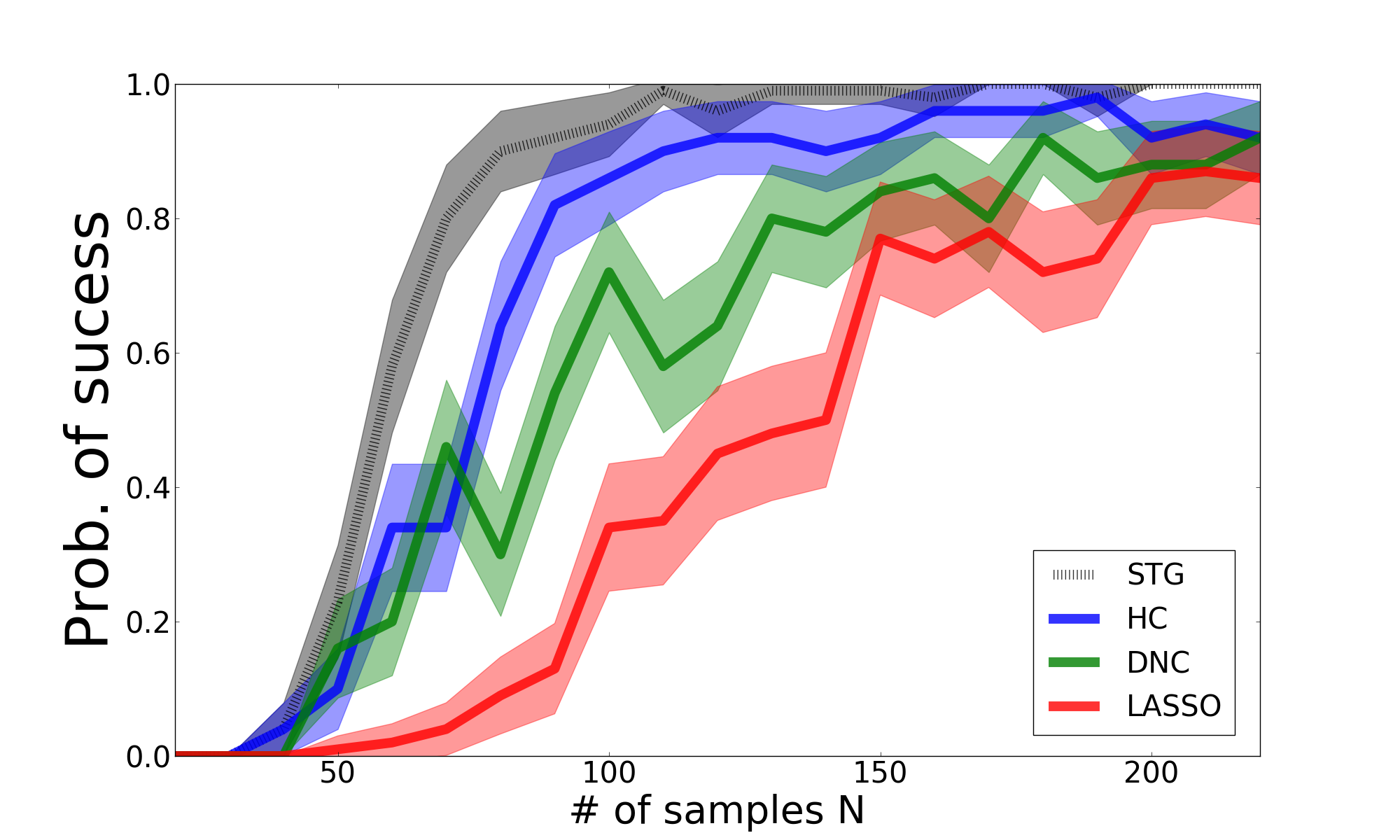}
\end{center}
\vskip -0.2in
\caption{Feature selection in linear regression using a correlated design matrix. Probability of recovering the informative features as a function of the number of samples. Comparison between STG, HC, LASSO and DNC.} \vskip -0.12in
\label{fig:correlated}
\end{figure}

Next we evaluate our method in a non-linear setting using a variant of the MADELON dataset, which includes correlated features.
%is more challenging than the one presented in Section \ref{sec:experiments}).
Following \cite{nips2003}, the first $5$ informative features of MADELON are used to create $15$ additional coordinates based on a random linear combination of the first $5$. A Gaussian noise $\mathcal{N}(0,1)$ is injected to each feature. Next, additional $480$ nuisance coordinates drawn from $\mathcal{N}(0,1)$ are added. Finally, $1\%$ of the labels are flipped. \footnote{generated using \texttt{dataset.make\_classification} from \texttt{scikit-learn} (\url{http://scikit-learn.org/})}
We use $1,500$ points from this dataset and evaluate the ability of STG to detect the informative features.

Fig. \ref{fig:madelon_lam} shows the precision of feature selection (black line) and the number of selected features (red line) as a function of the regularization parameter $\lambda$ in the range $[0.01,10]$. 
We observe that there is a wide range of $\lambda$ values in which our method selects only relevant features (i.e. the precision is $1$). Furthermore, there is a wide range of $\lambda$ values in which $5$ features are selected consistently.

 Next, we compare classification accuracy to Random Forest and LASSO using a $5$-fold cross validation. As evident from Fig. \ref{fig:madelon_acc}, STG achieves the highest accuracy while using less features. Moreover, as depicted from this figure, peak performance occurs when selecting $5$ features; thus, STG provides a clear indication to the true number of informative features. Both LASSO and RF on the other hand, do not provide a clear indication of the true number of relevant features.

\begin{figure}[htb!]
\vskip -0.16in
\begin{center}
\subfigure[]{\label{fig:madelon_lam}\includegraphics[width=0.46\textwidth] {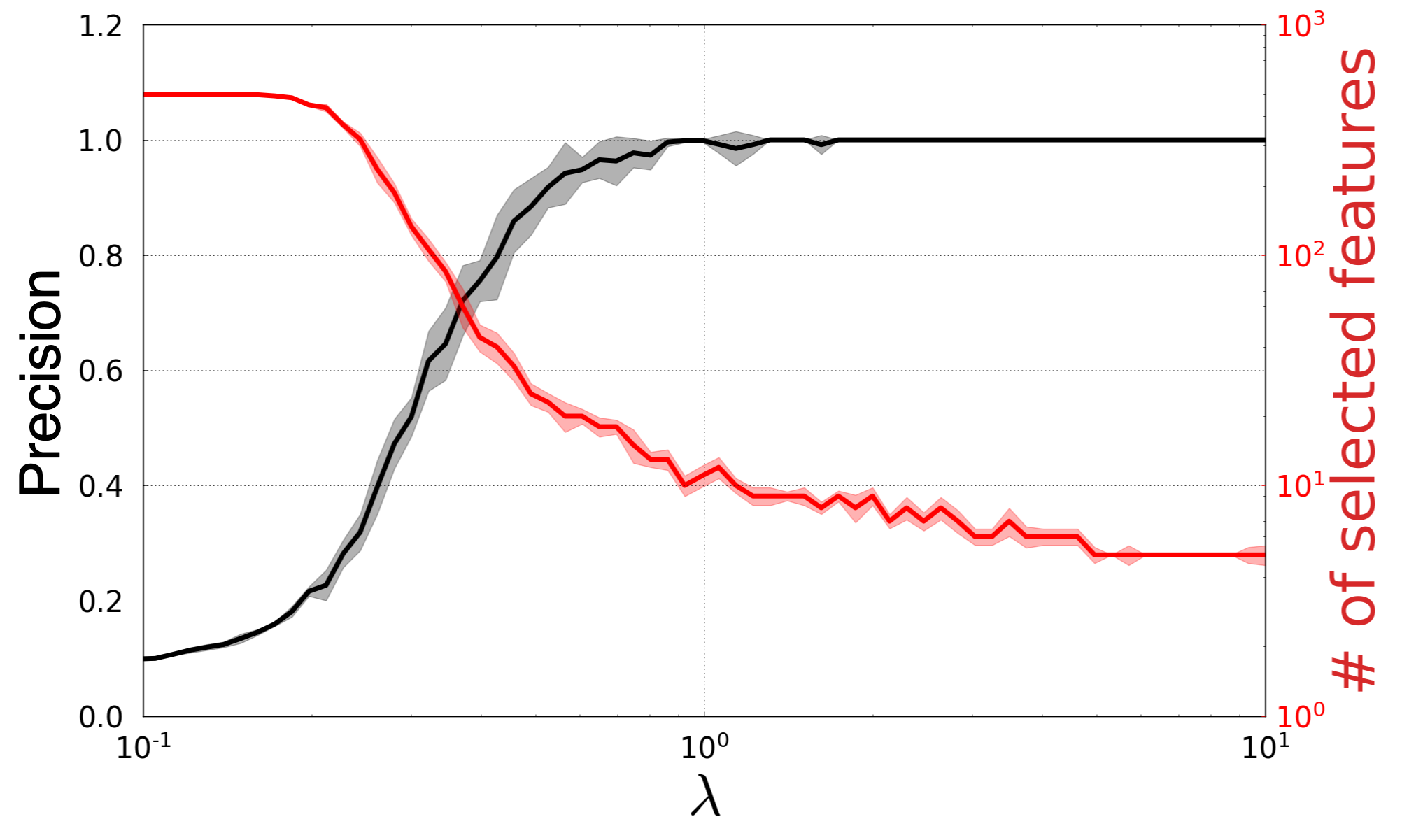}}
\subfigure[]{\label{fig:madelon_acc}\includegraphics[width=0.48\textwidth] {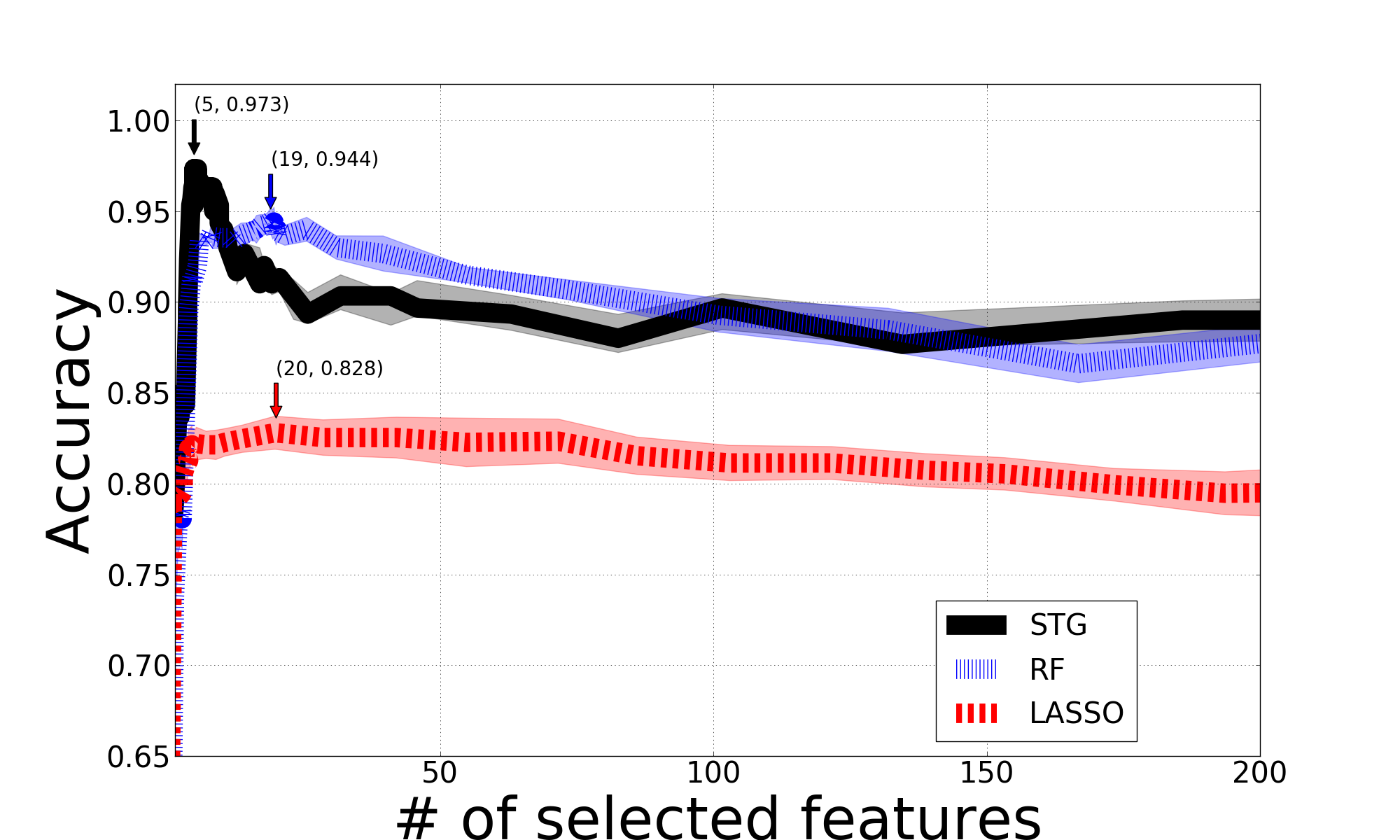}}
\end{center}
\vskip -0.22in
\caption{ \subref{fig:madelon_lam} An empirical evaluation of the effect the regularization parameter $\lambda$ has on the precision of feature selection (black line) and the number of selected features (red line). The precision and the number of selected features are presented on the left and right side of the $y$-axis, respectively. The means are displayed as solid lines while the standard deviations are marked as shaded regions around the means. \subref{fig:madelon_acc} Classification accuracy on the MADELON datasets. We evaluate performance using $5$-fold cross validation for different number of selected features. In this dataset, the first $5$ coordinates are strongly relevant features and the next $15$ are weakly relevant. In that regime the proposed method outperforms RF and LASSO.}
\vskip -0.18in
\end{figure}

}
\section{Conclusion}

In this paper, we propose a novel embedded feature selection method based on stochastic gates. 
It has an advantage over previous $\ell_1$ regularization based methods in its ability to achieve a high level of sparsity in nonlinear models such as neural networks, without hurting performance.

We justify our probabilistic feature selection framework from an information theoretic perspective. 
In experiments, we demonstrate that our method consistently outperforms existing embedded feature selection methods in both synthetic datasets and real datasets.

\section*{Acknowledgements}
The authors thank Nicolas Casey and the anonymous reviewers for their helpful feedback.
%This research was funded in part by NSF grant \textbf{number}, NSF , NIH grant \textbf{number}, and the Funai Overseas Scholarship to YY.
This work was supported by the National Institutes of Health [R01GM131642, R01HG008383, P50CA121974 and R61DA047037], National Science Foundation DMS 1723128, and the Funai Overseas Scholarship to YY.

\bibliography{ref}
\bibliographystyle{IEEEtran}
%\appendix
\clearpage

\beginsupplement
\section*{Supplemental Material}

\section{Proof of Proposition~\ref{prop:1}}
We now provide a proof for proposition \ref{prop:1} showing the equivalence between the stochastic optimization ~\eqref{eq:beropt} and the deterministic one \eqref{eq:MI_c}. We start by considering ${\cal{S}}'$ be a subset such that ${\cal{S}}^* \setminus {\cal{S}}' \neq \emptyset$. That is there exists some element in ${\cal{S}}^*$ that is not in ${\cal{S}}'$. For any such set ${{\cal{S}}}'$ we have that $I(\myvec{X}_{{\cal{S}'}}; Y)< I(\myvec{X} ; Y)$. Indeed, if we let $i \in {\cal{S}}^* \cap {{\cal{S}}}'^c$ then we have
\begin{align*}
    I(\myvec{X}_{\cal{S}'} ; Y) & \leq I(\myvec{X}_{\setminus \{i\}} ; Y) \\
    & = I(\myvec{X} ; Y) - I(\myvec{X}_i ; Y | \myvec{X}_{\setminus \{i\}}) \\
    & < I(\myvec{X} ; Y),
\end{align*}
where the final inequality follows by Assumption 1. On the other hand, for any subset ${\cal{S}'}$ such that $S^* \subset {S'}$, based on Assumption 2 we get that $I(\myvec{X}_{{S'}} ; Y) = I(\myvec{X} ; Y)$. Now, when we consider the Bernoulli optimization problem we have
\begin{equation*}
\max_{\myvec{\pi}} I(\myvec{X} \odot {\myvec{\tilde{S}}}; \myvec{Y})   \quad  \text{s.t.} \quad \sum_l \myvec{\pi}_l \le k \text{ and } 0 \leq \myvec{\pi}_l \leq 1.
\end{equation*}
The mutual information can be expanded as 
\begin{equation*}
    I(\myvec{X} \odot {\myvec{\tilde{S}}}; {Y}) = \sum_{\tilde{{s}}} I(\myvec{X} \odot \myvec{\tilde{s}} ; Y) p_{\myvec{\pi}}({\tilde{\myvec{S}}}=\myvec{\tilde{s}}),
\end{equation*}
where we have used the fact that $\myvec{\tilde{S}}$ is independent of everything else. Our goal is to understand the form of the distributing of the random vector $\myvec{\tilde{S}}$ that maximizes the objective subject to its constrains. Recall that in optimization~\eqref{eq:beropt} the coordinates of $\myvec{\tilde{S}}$ are sampled at random, which is motivated by practical needs. This means that the distribution that is being optimized over $p_{\myvec{\pi}}$ is a product distribution. Here, we will show that we can remove independence assumption. If we can show that the distribution found by solving the more general optimization problem is still a product distribution, then we obtain a solution to the original optimization~ \eqref{eq:beropt}. 

Now, from above we know that the optimal value of the optimization is $I(\myvec{X}_{\cal{S}'}; {Y})$ for any set ${\cal{S}}^* \subset \cal{S}'$. Hence, any unconstrained distribution should place all of its mass on such subsets in order to maximize the mutual information. As a result $\sum_{l \in S^*} p_{\myvec{\pi}}(\myvec{\tilde{S}}_l = 1) = k$. However, there is an optimization constraint that $\exval [\sum_l \myvec{\tilde{S}}_l ] \leq k$.
Therefore, $\exval[\myvec{\tilde{S}}_l] = 0$ for any $l \notin {\cal{S}}^*$. Hence, the optimal solution is to select the distribution so that all of the mass is placed on the subset $S^*$ and no mass elsewhere. As this is also a product distribution, this complete the proof of the claim.

\section{Bridging the Two Perspectives}
\label{sec:vlb}
To motivate the introduction of randomness into the risk, we have looked at the feature selection problem from a MI perspective. Based on the MI objective, we have observed that introducing randomness into the constrained maximization procedure does not change the objective (See Proposition 1 in Section \ref{sec:MI}) 
Here we provide a relation between the MI objective ~\eqref{eq:MI_c} and the empirical risk ~\eqref{eq:optim}, which supports our proposed procedure.

We first note that the MI maximization over the set $\cal{S}$ can be reformulated as the minimization of the conditional entropy $H(Y|\myvec{X}_{\cal{S}})$ since $H(Y)$ does not depend on $\cal{S}$:
\begin{equation*}
    \max_{\cal{S}} I(\myvec{X}_{\cal{S}}; Y) \iff \min_{\cal{S}} H(Y|\myvec{X}_{\cal{S}}).
\end{equation*}

Recall that $\myvec{X}_{\cal{S}} = \myvec{X} \odot \myvec{\tilde{S}}$. 
By Proposition 1, we rewrite the deterministic search over the set $\cal{S}$ by a search over the Bernoulli parameters $\myvec{\pi}$: 
\begin{align*}
    \min_{\myvec{\pi}} H(\myvec{Y}|\myvec{X} \odot \tilde{S} ) &= \min_{\myvec{\pi}} \mathbb{E}_{X,Y,\tilde{S}} \left [ -\log P_{\myvec{\theta}^*}(Y|\myvec{X} \odot \tilde{\myvec{S}}) \right ] \\
    &= \min_{\myvec{\pi},\myvec{\theta}} \mathbb{E}_{X,Y,\tilde{S}} \left [ -\log P_{\myvec{\theta}}(Y|\myvec{X} \odot \tilde{\myvec{S}}) \right ], 
\end{align*}
where the expectation is over $\myvec{X}, Y \sim P_{\myvec{\theta}*},$ which is the true data distribution, and $\tilde{\myvec{S}} \sim Bern(\tilde{\myvec{S}}|\myvec{\pi})$ and we put our model distribution as $P_{\myvec{\theta}}.$ Then we can rewrite the right hand side as:
\begin{align*}
\mathbb{E}_{X,Y,\tilde{S}}\log P_{\myvec{\theta}^*}(Y|\myvec{X} \odot \myvec{\tilde{S}}) &= \mathbb{E}_{X,Y,\tilde{S}}\left [ \log P_{\myvec{\theta}^*}(Y|\myvec{X} \odot \myvec{\tilde{S}}) \frac{P_{\myvec{\theta}}(Y|\myvec{X} \odot \myvec{\tilde{S}})}{P_{\myvec{\theta}}(Y|\myvec{X} \odot \myvec{\tilde{S}})} \right] \\
&= \mathbb{E}_{X,Y,\tilde{S}} \left [ \log \frac{P_{\myvec{\theta}*}(Y|\myvec{X} \odot \myvec{\tilde{S}})}{P_{\myvec{\theta}}(Y|\myvec{X} \odot \myvec{\tilde{S}})} \right ] + \mathbb{E}_{X,Y,\tilde{S}} \left [ \log P_{\myvec{\theta}}(Y|\myvec{X} \odot \myvec{\tilde{S}}) \right ].
\end{align*}

Since $\text{KL}(P_{\myvec{\theta}^*}(Y|\myvec{X} \odot \myvec{\tilde{S}}) || P_{\myvec{\theta}}(Y|\myvec{X} \odot \myvec{\tilde{S}}))$ is non-negative, $\mathbb{E}_{\tilde{S}} \text{KL}(P_{\myvec{\theta}*}(Y|\myvec{X} \odot \myvec{\tilde{S}}) || P_{\myvec{\theta}}(Y|\myvec{X} \odot \myvec{\tilde{S}}))$ is also non-negative because it is a weighted sum of non-negative terms.
Noting that 
\begin{equation*}
    \mathbb{E}_{\tilde{S}} \text{KL}(P_{\myvec{\theta}*}(Y|\myvec{X} \odot \myvec{\tilde{S}}) || P_{\myvec{\theta}}(Y|\myvec{X} \odot \myvec{\tilde{S}}) = \mathbb{E}_{X,Y,\tilde{S}} \left [ \log \frac{P_{\myvec{\theta}*}(Y|\myvec{X} \odot \myvec{\tilde{S}})}{P_{\myvec{\theta}}(Y|\myvec{X} \odot \myvec{\tilde{S}})} \right ]
\end{equation*}
we conclude that 
\begin{equation*}
    \mathbb{E}_{X,Y,\tilde{S}} \left [- \log P_{\myvec{\theta}^*} (Y|\myvec{X} \odot \myvec{\tilde{S}}) \right ] \le \mathbb{E}_{X,Y,\tilde{S}} \left [ -\log P_{\myvec{\theta}}(Y|\myvec{X} \odot \myvec{\tilde{S}}) \right ].
\end{equation*}
If we consider the negative log likelihood of the target given the observations (i.e. $-\log P_{\myvec{\theta}}(Y|\myvec{X} \odot \myvec{\tilde{S}})$) as a loss function $L$ (in Eq. \eqref{eq:optim}), then we see that the minimizing the risk approximately maximizes the MI objective in \eqref{eq:MI_c}.

\section{Details of the Regularization Term}

\begin{figure}[htb!]
\begin{center} 
\includegraphics[width=0.5\textwidth]{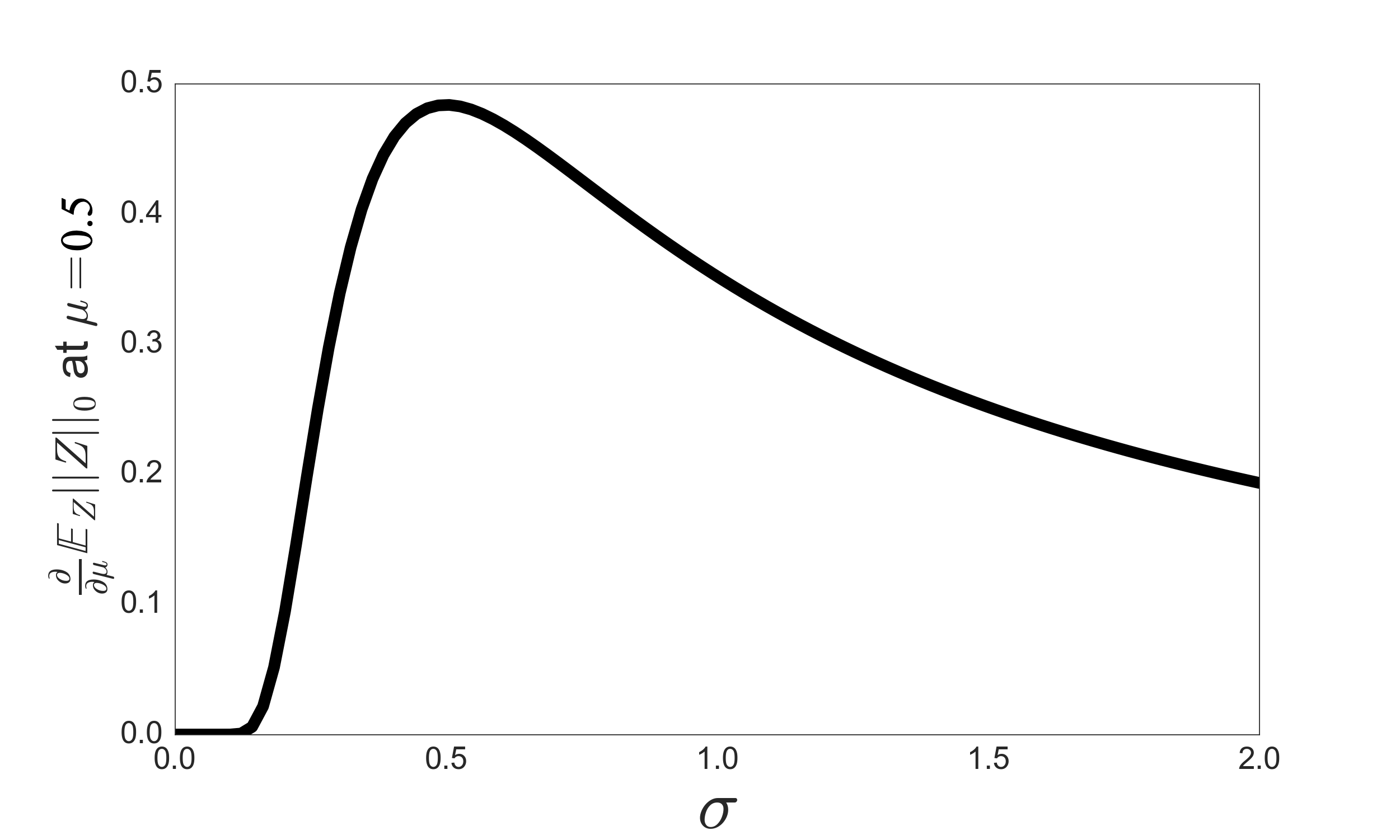}
\end{center}
\caption{The value of $\frac{\partial}{\partial \mu}\mathbb{E}_Z ||\myvec{Z}||_0 \vert _{\mu=0.5} = \frac{1}{\sqrt{2\pi \sigma^2}} e^{-{\frac{1}{8\sigma^2}}}$ for $\sigma = [0.001, 2]$.}
\label{fig:reg_grad}
\end{figure}

Here we provide a detailed description of our regularization term.
For the vector of stochastic gates $\myvec{z} \in \mathbb{R}^D$, the regularization term is expressed as follows: 
\begin{align*}
    \mathbb{E}_Z \norm{\myvec{Z}}_0 &= \sum_{d=1}^D \mathbb{P}[z_d > 0] = \sum_{d=1}^D \mathbb{P}[\mu_d + \sigma \epsilon_d > 0] \\
    &= \sum_{d=1}^D \{ 1 - \mathbb{P}[\mu_d + \sigma \epsilon_d \le 0] \} \\
    &= \sum_{d=1}^D \{ 1 - \Phi(\frac{-\mu_d}{\sigma}) \} \\
    &= \sum_{d=1}^D \Phi\left (\frac{\mu_d}{\sigma} \right) .
\end{align*}
Note that the derivative of the regularization term with respect to the distribution parameter $\mu_d$ is simply the Gaussian PDF:
\begin{align*}
    \frac{\partial}{\partial \mu_d} \mathbb{E}_Z \norm{\myvec{Z}}_0 &= \frac{\partial}{\partial \mu_d}\Phi(\frac{\mu_d}{\sigma})
    %-\frac{1}{2} \frac{2}{\sqrt{\pi}} e^{-t^2} \frac{\partial t}{\partial \mu_j}
    = \frac{1}{\sqrt{2\pi \sigma^2}} e^{-{\frac{\mu_d^2}{2\sigma^2}}}. 
\end{align*}
We now turn our attention towards providing a scheme for selecting $\sigma$
The effect of $\sigma$ can be understood by looking at the value of $\frac{\partial}{\partial \mu_d} \mathbb{E}_Z || \myvec{Z}||_0$. 
In the first training step, $\mu_d$ is 0.5. Therefore, at initial training phase, $\frac{\partial}{\partial \mu_d} \mathbb{E}_Z || \myvec{Z}||_0$ is close to $\frac{1}{\sqrt{2\pi \sigma_d^2}} e^{-{\frac{1}{8\sigma_d^2}}}$. In order to remove irrelevant features, this term (multiplied by the regularization parameter $\lambda$) has to be greater than the derivative of the loss with respect to $\mu_d$ because otherwise $\mu_d$ is updated in the incorrect direction. To encourage such behavior, we set $\sigma = 0.5$, which is around the maximum of the gradient during the initial phase as shown in Fig. \ref{fig:reg_grad}. Although the point that attains the maximum moves as $\mu$ changes, we empirically observe that setting $\sigma=0.5$ performs well in our experiments when the regularization parameter $\lambda$ is appropriately set.
Note that in our implementation we divide the regularization term by the size of the features $D$. This rescaling normalizes the hyper-parameter search into a similar range.
    
\section{Issues in Gradient Estimation of Discrete Random Variables}
\label{sec:ge}
In Section \ref{sec:cr}, we have introduced Bernoulli random variables $\tilde{s}_d,d=1,...,D$ with corresponding parameters $\pi_d$ into the risk objective \eqref{eq:bern_risk}. Taking the expectation over the $\ell_0$ norm of $\myvec{\tilde{S}}$ boils down to the sum of the Bernoulli parameters $\pi_d$. However, a gradient-based optimization over the resulting objective suffers from high variance due to the discrete nature of $\myvec{\tilde{S}}$. Here, we attempt to convey this problem by analyzing the risk term in the objective \eqref{eq:bern_risk}.

Using the Bernoulli paramterization the empirical risk $\hat{R}(\myvec{\theta},\myvec{\pi})$ is expressed as

\begin{align*}
\sum_{\myvec{z} : \{0,1\}^D} \left [ \sum_{n=1}^N \left [  L(f_{\theta}(\myvec{z} \odot \myvec{x}_n), \myvec{y}_n \right ] \prod_{d=1}^D  \pi_d^{z_d} (1 - \pi_d)^{1-z_d} \right ].
\end{align*}
In practice, as the outer sum involves enumerating $2^D$ possibilities of the indicator variables, one can replace the outer sum with Monte Carlo samples from the product of Bernoulli distributions $B(\myvec{z}|\myvec{\pi})$. However, a Monte Carlo estimate of $\frac{\partial}{\partial \pi_d} \hat{R}(\myvec{\theta}, \myvec{\pi})$ suffers from high variance. To see this, consider the following exact gradient of the empirical risk with respect to $\pi_d$, which is 
\begin{align*}
\sum_{\myvec{z}: \{0,1\}^D, z_d=1} \left [ L(\myvec{z}) p_{z_{i \neq d}} \right ]   - \sum_{\myvec{z}: \{0,1\}^D, z_d=0} \left [ L(\myvec{z}) p_{z_{i \neq d}} \right ],
\end{align*}
where $p(z_{i \neq d}) = \prod_{i \neq d}^D  \pi_i^{z_i} (1 - \pi_i)^{1-z_i}$, by absorbing the model $f_{\theta}(\cdot)$ and the data into $L(\cdot)$. 
Due to the discrete nature of $\myvec{z}$, we see that even the sign of the gradient estimate becomes inaccurate if we can only access a small number of Monte Carlo samples. While a score-function estimator such as REINFORCE \cite{REINFORCE} can be used, it is known that the reparametrization trick is more effective for variance reduction \cite{Gumbel1,Gumbel2,Louizos2017LearningSN}. 
%allows the model to access the gradient information $L(z)'$, which reduces the variance in practice. 

\section{Hard-Concrete Distribution (HC)}
In the main text, we have compared the proposed embedded method using our STG distribution and an alternative based on the Hard-Concrete (HC). 
Here, we provide a full description for the HC distribution. The HC was introduced in \cite{Louizos2017LearningSN} as a modification of Binary Concrete \cite{Gumbel1,Gumbel2}, whose sampling procedure is as follows:
\begin{align*}
u \sim U(0,1), L = \log(U) - \log(1-U), \\
s = \frac{1}{1 + \exp (\frac{- (\log \alpha + L)}{\beta})}, \\
\bar{s} = s (\zeta - \tau) + \tau, \\
z = \min (1, \max(0, \bar{s}) ), 
\end{align*}
where $(\tau, \zeta)$ is an interval, with
$\tau <0$  and $\zeta >1$. 
This induces a new distribution, whose support is  $[0, 1 ]$ instead of $(0, 1)$. With $0 < \beta < 1$, the
probability density concentrates its mass near the 
end points, since values larger than $\frac{1 - \tau}{\zeta - \tau}$ are rounded to one, whereas 
values smaller than $\frac{-\tau}{\zeta - \tau}$ are 
rounded to zero. 

The CDF of $s$ is 
\begin{equation}
 Q_{s}(s | \beta, \log \alpha) = \text{Sigmoid}( (\log s - \log (1 -s ) ) \beta - \log \alpha),  
\end{equation}

and so the CDF of $\bar{s}$ is
\begin{equation}
 Q_{\bar{s}}(\bar{s} | \phi) = \text{Sigmoid}( (\log (\frac{\bar{s} - \tau}{\zeta - \tau} ) - \log (1 - \frac{\bar{s} - \tau}{\zeta - \tau} ) ) \beta - \log \alpha ), 
\end{equation}
where $\phi = (\beta, \log \alpha, \zeta, \tau) $.
Using this distribution to model the gates, the probability of a gate $z$ being active is $1 - Q_{\bar{s}}(0 | \phi)$ and can be written as 
\begin{align}
1 - Q_{\bar{s}}(0|\phi)= \text{Sigmoid}( \log \alpha - \beta \log \frac{- \tau}{\zeta} ) .
\end{align}
The hyperparameters $\beta,\xi$ and $\tau$ are set as in \cite{Louizos2017LearningSN} and fixed throughout training, while $\alpha$ is learned to determine whether the gate is active or not.
Note that $L$ is distributed as a Logistic distribution, which causes high variance in $\bar{s}$ due to the heavy-tailness of the distribution.
By replacing the logistic distribution with the Gaussian distribution, we reduce variance in gradient estimates, leading to stability of feature selection (See Section \ref{sec:discussion}).

\section{Additional Experiments}

\subsection{Two Moons classification with nuisance features} \label{sec:twomoon}
\begin{figure}[htb!]
\begin{center}
\subfigure[]{\label{fig:twomoons}\includegraphics[width=0.45\textwidth] {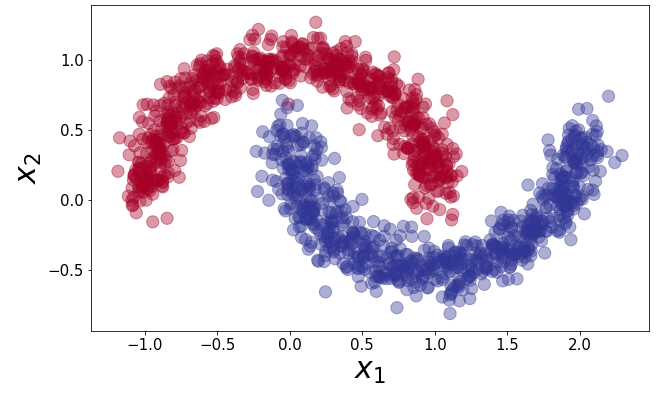}}
\subfigure[]{\label{fig:classacc1}\includegraphics[width=0.45\textwidth] {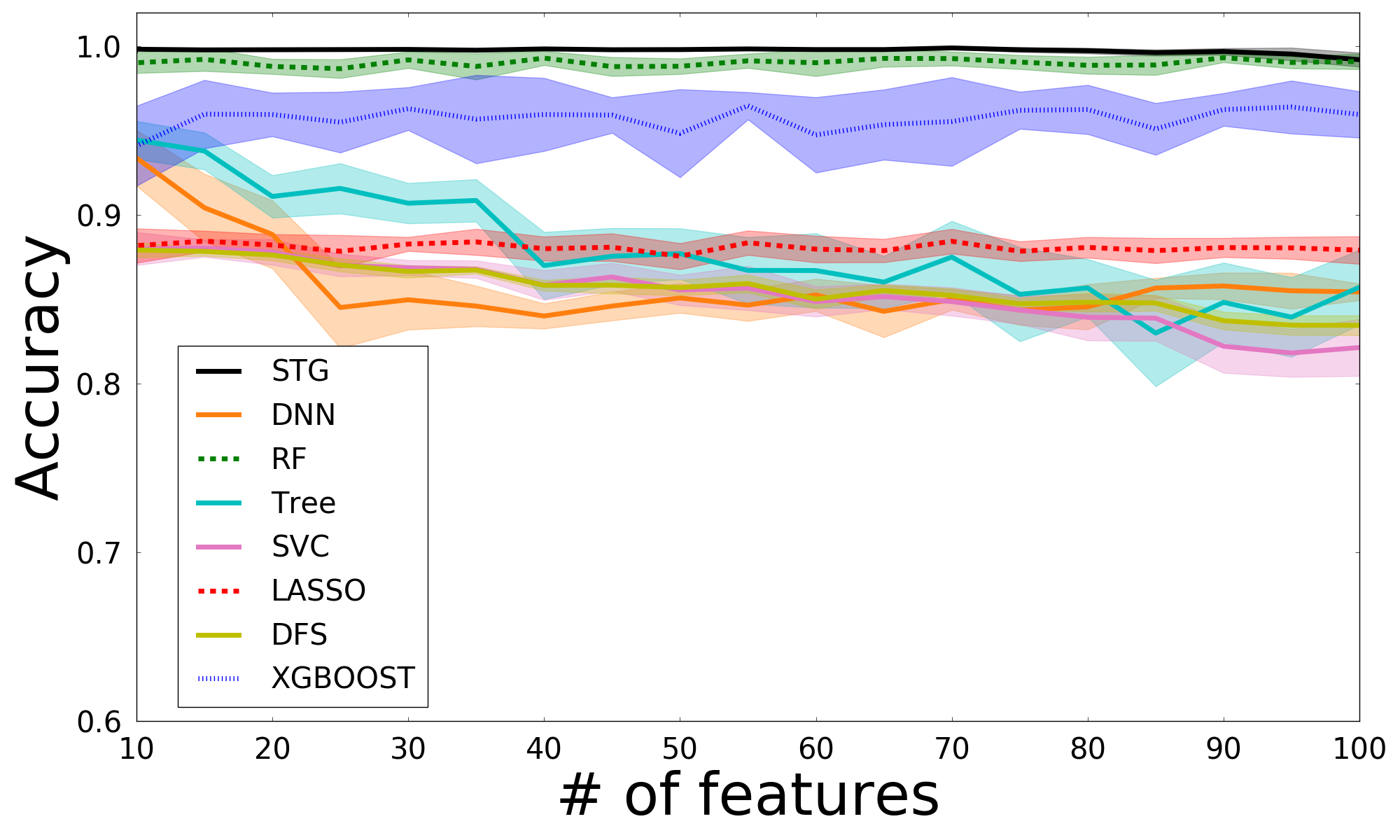}}

\end{center}
\caption{\subref{fig:twomoons} Realizations from the "Two moons" shaped binary classification class. $x_1$ and $x_2$ are the relevant features, $x_i,i=3,...,D$ are noisy features drawn from a Gaussian with zero mean and variance of $1$. \subref{fig:classacc1} Classification accuracy (mean and standard deviation based on 20 runs) vs. the number of irrelevant noisy dimension.} 
\end{figure}

In this experiment, we construct a dataset based on "two moons" shape classes, concatenated with noisy features. The first two coordinates  $x_1, x_2$ are generated by adding a Gaussian noise with zero mean and the variance of $\sigma^2_r=0.1$ onto two nested half circles, as presented in Fig. \ref{fig:twomoons}. Nuisance features $x_i,i=3,...,D$, are drawn from a Gaussian distribution with zero mean and variance of $\sigma_n^2=1$. We reserve the $70\%$ as a test set, and use $10\%$ of the remaining training set as a validation set. We follow the same hyperparameter tuning procedure as in the XOR experiment.
The classification accuracy is in Fig. \ref{fig:classacc1}. Based on the classification accuracies, it is evident that for a small number of nuisance dimensions all methods correctly identify the most relevant features. The proposed method (STG) and Random Forest (RF) are the only methods that achieve near perfect classification accuracy for a wide range of nuisance dimensions. 
The other NN based method (DFS) seem to converge to sub-optimal solutions. We note that all the methods achieve the median rank 1.5, which is the optimal median rank  in this example. 

\subsection{Convergence Comparison to Hard-Concrete distribution}

In this section, we show additional experiments to compare STG with HC in terms of convergence speed.

\begin{figure}[htb!] 
\begin{center} 
\subfigure[]{\label{fig:ifwra}\includegraphics[width=0.45\textwidth]{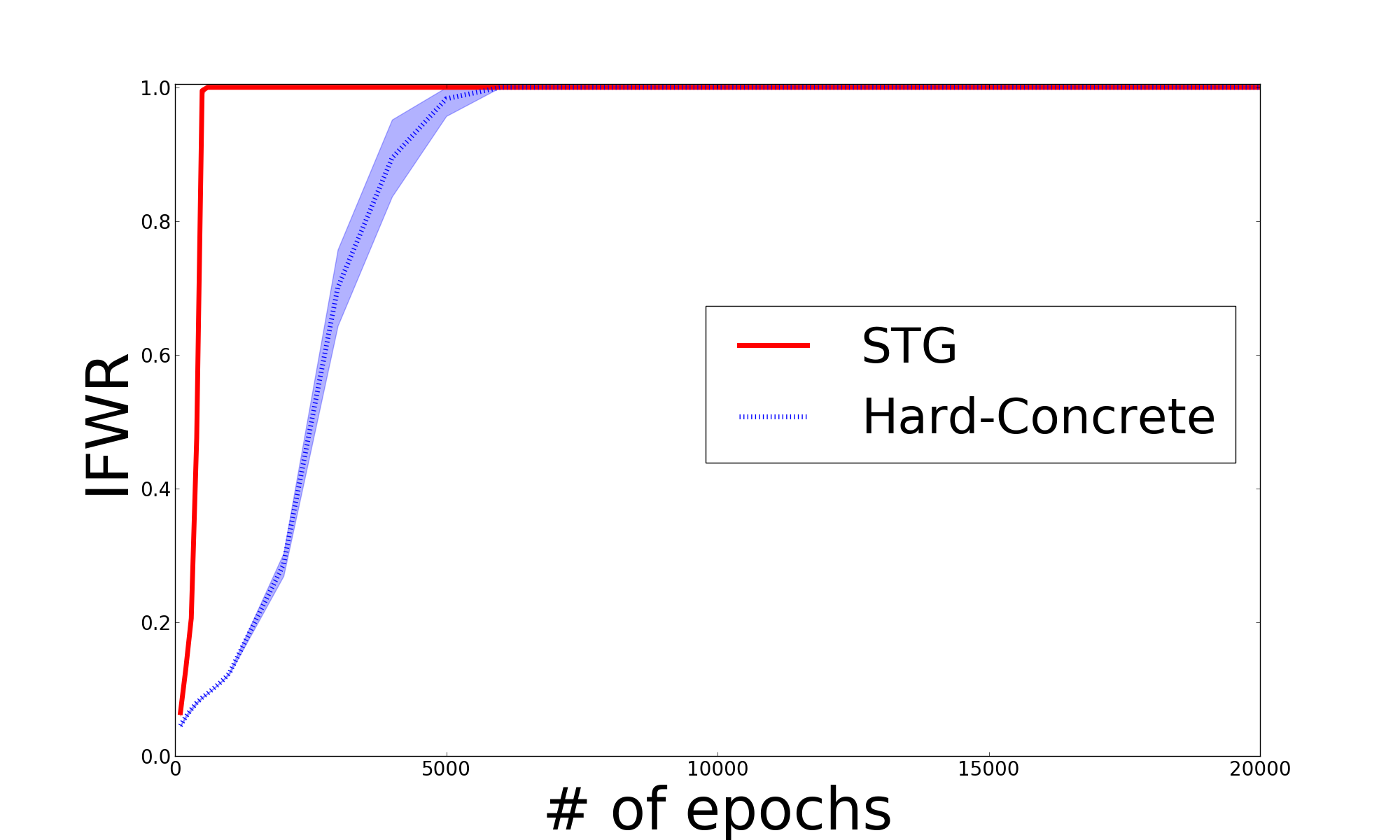}}
\subfigure[]{\label{fig:ifwrb}\includegraphics[width=0.45\textwidth]{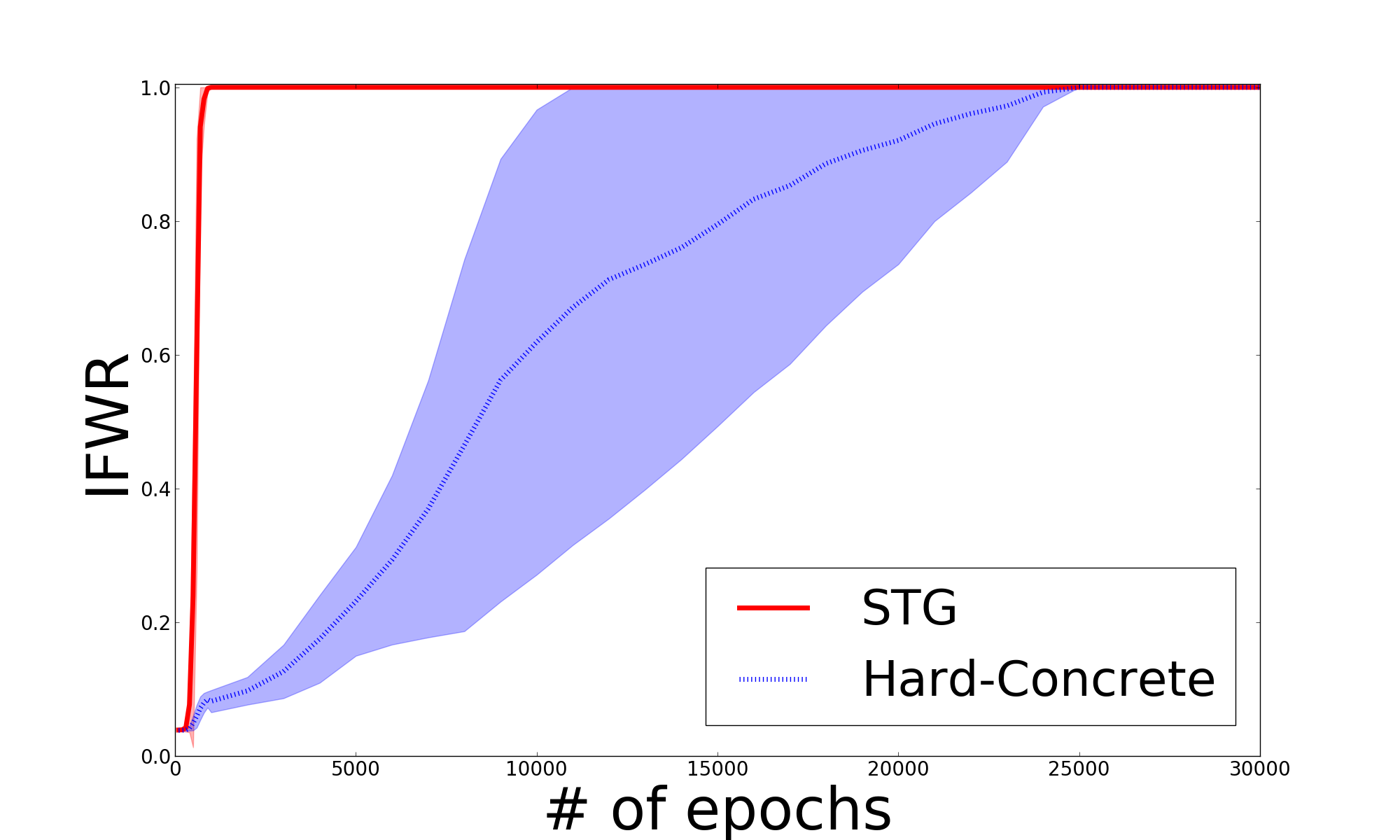}}
\subfigure[]{\label{fig:ifwrc}\includegraphics[width=0.45\textwidth]{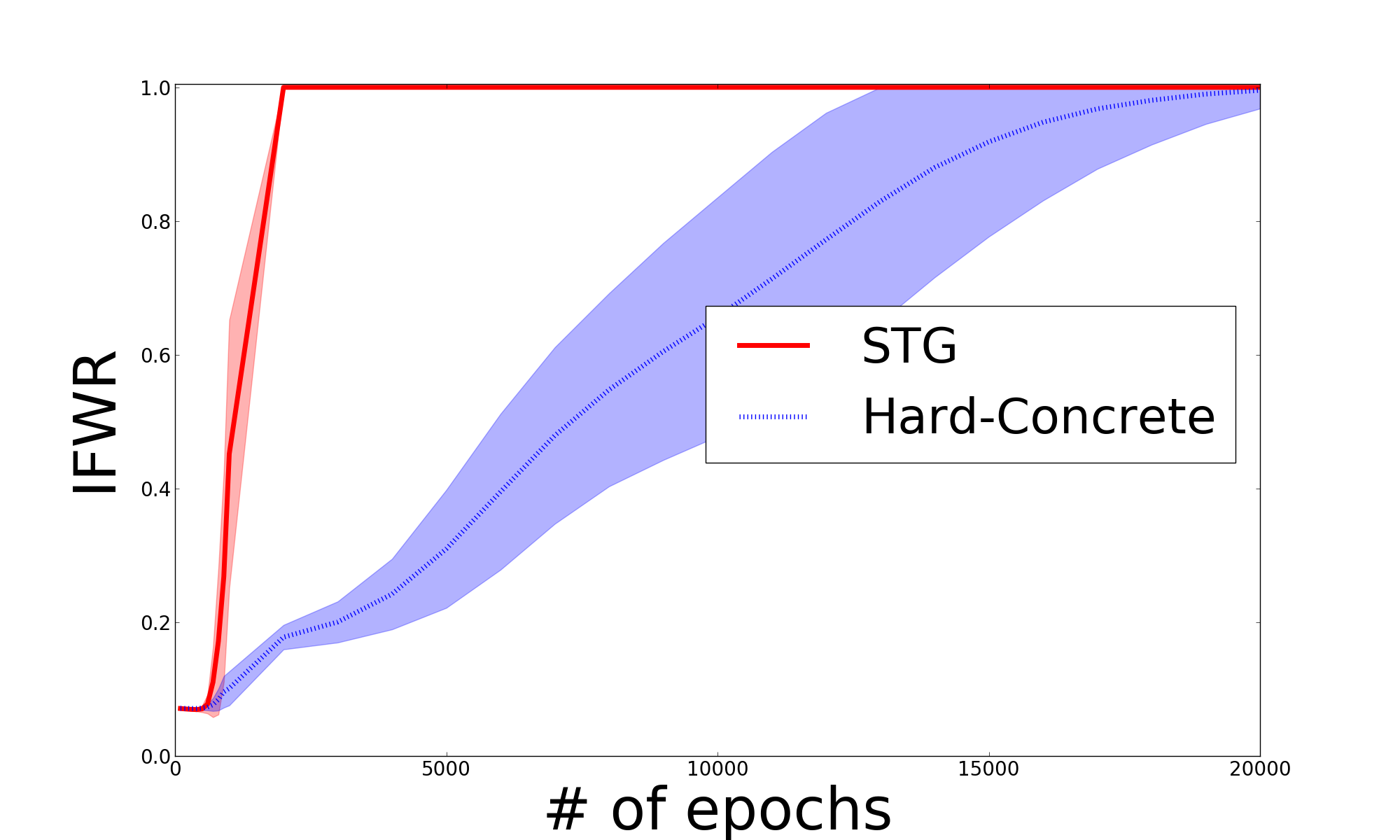}}
\end{center}
\caption{Comparison between STG and Hard-Concrete on the Two-Moon dataset \subref{fig:ifwra}, XOR \subref{fig:ifwrb} and MADELON \subref{fig:ifwrc}. The shaded area represents the standard deviation, calculated by running the same experiment 10 times with different random initializations.}
\label{fig:hardconcrete}
\end{figure}

The main difference between our proposed distribution (STG) and the Hard-Concrete \cite{Louizos2017LearningSN} distribution is that the latter is based on the logistic distribution,
which has a heavier tail than the Gaussian distribution we have employed.  
As shown in Fig \ref{fig:hardconcrete}, the heavy-tailness results in instability during training. Furthermore, the STG converges much faster and more reliably than the feature selection method using the HC distribution on a the two-moons, XOR and MADELON datasets (see Subsection \ref{sec:synth} and \ref{sec:twomoon}).

\section{Reproducibility}
Here we provide a full description of the procedures and datasets we have used in the experimental parts of the paper.
\subsection{Datasets}

\subsection{Reuters Corpus Volume I }
The Reuters Corpus Volume I (RCV1) consists of $800,000$ news stories manually labeled by $103$ categories. 
This is a multilabel regime, where each story is assigned to multiple categories. 
Here we focus on a binary subset of this corpus, with $23,203$ stories, where we use $10 \%, 8.5\%, \text{and } 81.5\%$ for the train, validation and test set, respectively.
The total number of feature is $47,236$. This example demonstrates that our method is also effective in an extremely high dimensional regime of nonlinear function estimation.

\subsection{Purified populations of peripheral
blood monocytes (PBMCs)} \label{sec:pbmc}

Single-cell RNA sequencing (scRNA-seq) is a novel technology that measures gene expression levels of hundreds of thousands of individual cells. 
\cite{pbmcP}, have subjected more than $90,000$ purified populations of peripheral
blood monocytes (PBMCs) to scRNA-seq analysis. Here we focus on classifying two subpopulations of T-cells, namely the Naive and regulatory T-cells. 

The data consists of $D=11,382$ genes and $N=20742$ cells, of which we only use $10\%$ of the data for training. We apply the proposed method for different values of $\lambda$ and report the number of selected features and classification accuracy on the test set. In this example, the STG identifies $20-30$ genes which are sufficient for the classification task.

\subsection{Details of Cox Proportional Hazard Model}

Survival times are assumed to follow a distribution, which
is characterized by the survival function $S(t) = P(T >t)$.
A hazard function, which measures the instantaneous rate of death, is defined by $h(t) = \lim_{\Delta t \rightarrow 0} \frac{P(t < T  \le t + \Delta t | T>t)}{\Delta t} = \frac{p(t)}{S(t)}$.
We can relate the two functions in the following way: $S(t) = e^{- \int_0^t h(t) dt}$.

Proportional hazard models assume a multiplicative effect of the covariates $x$ on the hazard function such that $h(t|\myvec{x}) = h_0 (t) e^{\myvec{\theta}^T \myvec{x}}$,
where $h_0(t)$ is a baseline hazard function, which
is often the exponential or Weibull distribution, and $\myvec{\theta}$
is the parameter of interests. 

One of the difficulties in estimating $\myvec{\theta}$ in survival analysis is that a large portion of the available data is censored.
However, in order to obtain estimates, Cox observed that it is sufficient to maximize the partial-likelihood, which is defined as follows:
\begin{align*}
    L(\myvec{\theta}) = \prod_{T_i \text{uncensored}} \frac{e^{\myvec{\theta}^T \myvec{x}_i}}{\sum_{T_j \ge T_i} e^{\myvec{\theta}^T \myvec{x}_j}}.
\end{align*}

In \cite{Jared}, the authors propose DeepSurv, which uses a deep neural network model to replace the linear relations between the covariate $\myvec{x}$ and $\myvec{\theta}$, demonstrating improvements of survival time prediction over existing models such as CPH and the random survival forest \cite{Ishwaran07randomsurvival}, \cite{RSF}.

The Molecular Taxonomy of Breast Cancer International Consortium (METABRIC) dataset consists of gene expression data and clinical features for $1,980$ patients, and $57.72\%$ have an observed death due to breast cancer with a median survival time of 116 months.

The METABRIC dataset involves $24,368$ features (genes). Most genes are irrelevant for outcome prediction. To demonstrate the advantage of STG in the context of survival analysis, we selected 16 well-known genes relevant for survival (out of the 24,368 genes) that correspond to the Oncotype DX test, a gene panel used for treatment decision making. We also include five additional clinical features (hormone treatment indicator, radiotherapy indicator, chemotherapy indicator, ER-positive indicator,
and age at diagnosis). We then added 200 additional irrelevant gene variables that we selected randomly from the remaining list of genes. 

After we omit the null values, the number of samples is $1969$. We reserve the $20\%$ for test, and use the $20\%$ of the remaining training set as validation (that is, train: $1260$, valid: $315$, test: $394$ samples).

\paragraph{Experimental Detail}
For DeepSurv, we manually select the architecture using the validation set so that we obtain a similar performance reported in \cite{Jared}. The learning rate decay is set to 1.
The learning rate and the regularization parameter are optimized via Optuna \cite{OPTUNA} using the validation set, where the search range is set$ [1e-3, 1]$ for the learning rate and $ [1e-3, 1]$ for $\lambda$.
The hyperparameters used in the experiment are the following: architecture :[60, 20, 3], activation: Selu (as suggested by \cite{Jared}), learning rate : 0.152, learning rate decay: $1.0$, $\sigma$ : $0.5$, $\lambda: 0.023$, training epoch: 2000. 
Note that to see the effect of feature selection, we used the hyperparameters optimized for DeepSurv to test our method (Cox-STG).

\subsection{Implementation details} \label{sec:implement}
Datasets are first split into train, validation and test. Validation is always $10\%$ of the train, while the exact ratios between train and test is detailed for each experiment separately (see Table \ref{tab:realdata}).
All the neural network weights are initialized by drawing from $\mathcal{N}(0,0.1)$ and bias terms are set to $0$ (following Xavier initialization \cite{glorot2010understanding}). We use SGD for all the experiments, except for the Cox model where we use Adam.
All the experiments are conducted using Intel(R) Xeon(R) CPU E5-2620 v3 @2.4Ghz x2 (12 cores total). We set the number of Monte Carlo samples $K=1$, which worked well in our experiments. Hyper-parameters for all method are tuned using Optuna \cite{OPTUNA}. Optuna is a hyper-parameter optimization software that supports pruning and parallel computing across GPUs. We run $n=1000$ trails with parameters search ranges as described in Table \ref{table:range}.

\begin{table}[h]
\caption{List of the search range for the hyperparameters used in our expirements}
\begin{center}

\begin{tabular}{ |c||c| } 

 \hline
 Param  & Search range   \\ 
 \hline 
 \# dense layers & [1,5] \\
 \# hidden units & [10, 500]\\
 activation & [Tanh, Relu, Sigmoid] \\
 LR & [1e-4, 1e-1] \\ 
 n-epoch (DFS, SG-NN) & [50, 20000] \\
 $\alpha$ (SG-NN) & [1e-3, 10] \\
 $\lambda$ (SG-NN) & [1e-7, 10] \\
 $\lambda$ (STG, DFS) & [1e-3, 10]\\
 $\lambda$ (LASSO) & [1e-5, 1] \\ 
  n-est (RF, XGB, ERT) &  [5,100]    \\
n-boost-round (XGB) &  [1,100]    \\
  Thresh (RF, XGB, ERT) &  [1e-7,1]   \\
  max-depth (XGB) &  [1,5]    \\
  c (SVC) & [1e-7, 1]  \\
 \hline
 
\end{tabular}

\label{table:range}
\end{center}
\end{table}

For linear regression experiment, we use $0.1$ as a learning rate. For the XOR problem, the exact architectures used for the NN based methods are: (STG/HC): $[476, 490, 14]$ with Tanh, (DFS): $[100, 10]$ with Tanh, (SG-NN): $[100, 10, 5]$ with Tanh. For the two moons we use (STG): $[490, 406, 18]$ with Tanh, (DFS): $[158, 27, 224]$ with Tanh, (SG-NN): $[88, 28, 27]$ with Tanh. For the XOR problem, we attempted to use Optuna to optimize parameters of DFS and SG-L1-NN, but we ended up using the architecture suggested by the authors \cite{DFS, sparseNN-group-lasso} as they outperform the values suggested by Optuna. The number of epochs used for the XOR problem is $20k,14k,800$ for STG/HC, DFS and SG-L1-NN respectively. Regularization parameters are $0.17,3.3e-5$ and $3e-5$ respectively. The number of epochs used for the two-moons problem is $20K,1570,708$ for STG/HC, DFS and SG-NN respectively. Regularization parameters are $0.48,9e-3$ and $1e-3$ respectively. We note that the regularization parameters and learning procedure is different in nature, as we use an $\ell_0$ type penalty. For the PBMC experiment, the architecture use us $[27,10,383]$ with Tanh activations, a learning rate was $0.0036$, batch size is $1000$ and the number of epochs $8000$. The hyperparameter $\lambda$ varies in the range $[0.001,0.11]$ to achieve different levels of sparsity. 
For MADELON, GISETTE, ISOLET, we use the architecture optimized for the binary XOR classification. The number of epochs used are $20K,20K,4k$, the learning rate are $0.06,0.2,0.1$, batch size are $200,1000,40$. The parameters used are the following: (SE1) architecture $[600, 200, 100, 50]$ with ReLu activations, number of epochs is $5$, $\lambda: 5$, learning rate $ 0.0001$ (SE2) architecture $[600, 300, 150, 60, 20]$ with ReLu activations, number of epochs is $2000$, $\lambda: 5$, learning rate is $ 0.001$ (SE3) architecture $[600, 300, 150, 60$, 20] with ReLu activations, num epochs : $1000$, $\lambda: 1$, learning rate $ 0.005$. (RCP) architecture $[1000, 300, 150, 60, 20]$ with ReLu activation, num-epochs: $2000$, $\lambda: 5.0$, learning rate $ 0.001$. (REL) architecture $[26,91,63]$ with ReLu activation, number of epochs: $1600$, $\lambda: 0.031$, learning rate $ 0.007$. (RAI) architecture $[10,177]$ with ReLu activation, number of epochs: $1800$, $\lambda: 0.019$, learning rate is $ 0.002$. Architectures for SE1-SE3 and RCP where tuned manually. The ratio of train/test/valid split is 1:1:1 for synthetic regression data. For the real regression data (RCP and REL), the train size is $6000$, the test and valid size is $1000$ samples. For RAI the train size is $5000$, the test and valid size is $1000$ samples. For the Friedman data we use Tanh with an architecture of $[500,200,100,20]$ (HC/STG/DFS/SG-NN) with Tanh activations, a learning rate of $0.2$ and a batch size of 200. For BASEHOCK (STG/HC) the architechture used is $[500,105,25]$ with Tanh activations, a batch size of $50$, learning rate of $0.5$ and $2000$ epochs. For RELATHE (STG/HC) the architechture used is $[500,200,10]$ with Tanh activations, a batch size of $40$, learning rate of $0.008$ and $10k$ epochs. For COIL20 (STG/HC) the architechture used is $[32,438,158, 20]$ with Tanh activations, a batch size of $150$, learning rate of $0.12$ and $10k$ epochs. For PCMAC (STG/HC) the architechture used is $[109, 25,455]$ with Tanh activations, a batch size of $450$, learning rate of $0.5$ and $6000$ epochs. For RCV1 (STG/HC) the architechture used is $[500,100]$ with Tanh activations, a batch size of $500$, learning rate of $0.5$ and $650$ epochs. For MNIST (STG/HC) the architechture used is $[300,100]$ with Relu activations (as in \cite{mnist1}), a batch size of $200$, learning rate of $0.1$ and $250$ epochs.

In any experiment where we count the number of active features, we evaluate the set of indices such that: $\{d : \min(1, \max(0, \mu_d + 0.5)) > 0\}$ after training. In order to define the IFWR, for STG, the $d^{th}$ feature weight is set to $\max(0, \min(1, \mu_d + 0.5))$. For other neural net based methods, it is given by $\sum_j W_{dj}$, where $W$ is the weight matrix of the first layer. For other methods we just used the feature relevance returned by the trained model. Finally, the LASSO's IFWR in the XOR experiment was omitted from the manuscript as it suffered from high variance.

Regarding the comparison performed in the two-moons and XOR problem, we believe that adding IFWR along with classification accuracy versus number of feature selected provides a complementary perspective in demonstrating the efficacy of feature selection techniques. We emphasize that our goal is not to just rank features but select features by assigning the weight of $0$ to irrelevant features while simultaneously obtaining good predictive accuracy.

\end{document}